\documentclass[10pt,twocolumn,letterpaper]{article}

\usepackage{iccv}
\usepackage{times}
\usepackage{epsfig}
\usepackage{graphicx}
\usepackage{amsmath}
\usepackage{amssymb}
\usepackage[accsupp]{axessibility}  

\usepackage[utf8x]{inputenc} 
\usepackage[T1]{fontenc}    
\usepackage{url}            
\usepackage{booktabs}       
\usepackage{amsfonts}       
\usepackage{nicefrac}       
\usepackage{microtype}      
\usepackage{graphicx}
\usepackage{tikz}
\usepackage{tkz-euclide}
\usetikzlibrary{calc}
\usetikzlibrary{math}
\usetikzlibrary{shapes}
\usetikzlibrary{decorations.pathreplacing}

\usepackage{xcolor}
\definecolor{darkest}{RGB}{51, 0, 0}
\definecolor{dark}{RGB}{154, 0, 0}
\definecolor{light}{RGB}{255, 72, 72}
\definecolor{pale}{RGB}{255, 154, 154}

\usepackage[breaklinks=true,bookmarks=false]{hyperref}

\iccvfinalcopy 

\def\iccvPaperID{7559} 

\ificcvfinal\fi

\begin{document}

\title{DeepGaze IIE: Calibrated prediction in and out-of-domain\\for state-of-the-art saliency modeling}


\author{
Akis Linardos\footnotemark \\
University of Barcelona \\
{\tt\small linardos.akis@gmail.com} \\
\and
Matthias Kümmerer$^*$ \\
University of Tübingen \\
{\tt\small matthias.kuemmerer@bethgelab.org}
\and
Ori Press \\
University of Tübingen \\
{\tt\small ori.press@bethgelab.org} \\
\and
Matthias Bethge \\
University of Tübingen \\
{\tt\small matthias@bethgelab.org} \\

}

\maketitle

\ificcvfinal\thispagestyle{empty}\fi

\begin{abstract}
Since 2014 transfer learning has become the key driver for the improvement of spatial saliency prediction---however, with stagnant progress in the last 3-5 years. We conduct a large-scale transfer learning study which tests different ImageNet backbones, always using the same read out architecture and learning protocol adopted from DeepGaze II. By replacing the VGG19 backbone of DeepGaze II with ResNet50 features we improve the performance on saliency prediction from 78\% to 85\%. However, as we continue to test better ImageNet models as backbones---such as EfficientNetB5---we observe no additional improvement on saliency prediction.
By analyzing the backbones further, we find that generalization to other datasets differs substantially, with models being consistently overconfident in their fixation predictions. We show that by combining multiple backbones in a principled manner a good confidence calibration on unseen datasets can be achieved.
This new model ``DeepGaze IIE'' yields a significant leap in benchmark performance in and out-of-domain with a 15 percent point improvement over DeepGaze II to 93\% on MIT1003, marking a new state of the art on the MIT/Tuebingen Saliency Benchmark in all available metrics (AUC: 88.3\%, sAUC: 79.4\%, CC: 82.4\%).

\end{abstract}


\section{Introduction}
\renewcommand{\thefootnote}{\fnsymbol{footnote}}
\footnotetext{* indicates joint first authorship}
Saliency detection is involved in many sensory modalities. It summarizes the associated mechanisms as the ability of humans and animals to allocate their attention to the most important subsets of the data. In vision, this means attending to the elements of a visual input that stand out from their neighbouring regions, and visual saliency is usually operationalized by measuring fixations locations. Accordingly, in computer vision, saliency prediction currently refers to either predicting fixation locations or detecting salient objects.

Early on, researchers found out that the locations of fixations are statistically influenced by features of the visual stimuli that include both high-level properties such as people \cite{yarbus1967eye} and low-level ones such as spatial contrast \cite{contrast2016}. Soon after the \textit{Feature Integration Theory} emerged \cite{featureIntegration}, Koch and Ullmann outlined a computational mechanism to model attention \cite{koch1987shifts} which was implemented thirteen years later by Itti et al \cite{itti1998model}.
  
\begin{figure}[bp]
\begin{center}
\includegraphics[width=0.9\linewidth]{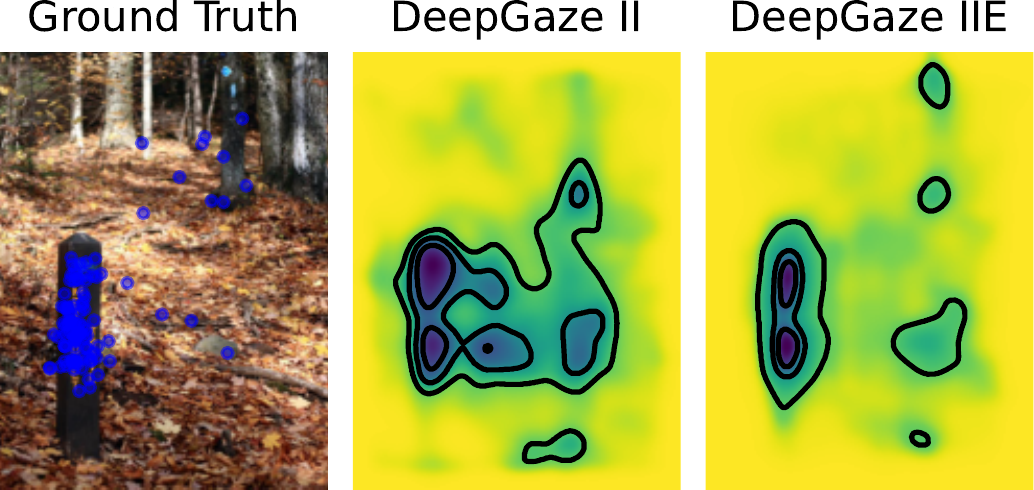}

\caption{By leveraging the diversity of different backbones, our new saliency model DeepGaze IIE very is able to predict human fixation locations very accurately.}
\label{fig:teaser}
\end{center}
\end{figure}
  
The Itti-Koch model was the first to predict a saliency map from any arbitrary image without the need to precompute elementary features allowing for a wide range of applications. This paved the way for many interesting saliency prediction models \cite{AIM, Kienzle, SUN} leading up to the present day where deep learning models are dominating the field \cite{eDN, deepgazei, kummererUnderstandingLowHighLevel2017, pan2017salgan, sotaEML, MSI-NET} driven by large scale saliency datasets \cite{MIT1003, SALICON, CAT2000}. As the saliency domain has substantially less data compared to some of the more prominent computer vision tasks, transfer learning has become the key driver for the improvement of spatial saliency prediction. One of the earliest works on transfer learning for deep learning is DeCAF \cite{DECAF}, where the authors used features extracted by a deep CNN that was trained on object recognition, leveraging a large dataset to tackle generic tasks that lacked labeled data. Following this transfer learning scheme, they outperformed the state-of-the-art on various vision challenges. Inspired by the huge success of deep convolutional models in the domain of classification and particularly the ImageNet benchmark \cite{ImageNet}, DeepGaze I \cite{deepgazei} was the first to transfer ImageNet learned features to the domain of saliency. Since then all high-performing saliency models use ImageNet as pretext task. 

\begin{figure*}[htbp]
\begin{center}
\includegraphics[width=0.9\textwidth]{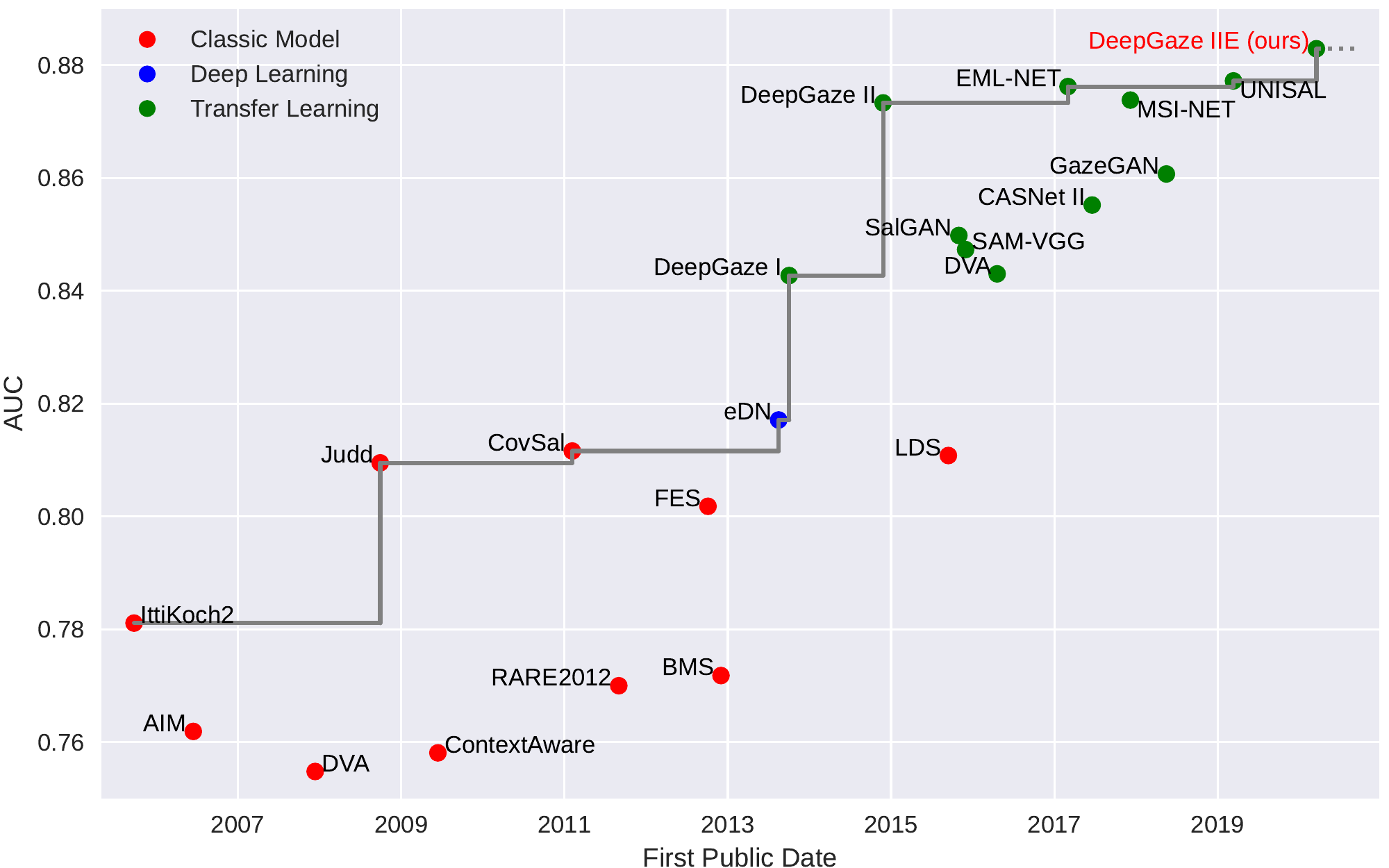}
\caption{A visualization of progress in saliency prediction over the last 15 years. The displayed dates correspond to the earliest date we found the models available, and usually reflect their first date the model was tested or the date of publication (whichever came first). The AUC corresponds to MIT300 evaluation at the MIT/Tuebingen saliency benchmark \cite{mit-tuebingen-saliency-benchmark}. For readability purposes we limited the scale of the plot to models whose AUC score is above 0.75. The gray line indicates state-of-the-art performance with respect to the models listed in the MIT/Tuebingen Saliency Benchmark. We could not include models which are only evaluated on the predecessor benchmark saliency.mit.edu since the evaluation changed slightly, resulting in different model scores.
}
\label{fig:over-the-years}
\end{center}
\end{figure*}

To this day, the problem of spatial saliency is far from solved and the simple case of the MIT300 benchmark \cite{BenchmarkMIT300} illustrates a substantial gap between existing models and the lower bound on the explainable information
(e.g., IG of 0.951 vs 1.317 and sAUC of 0.784 vs 0.823). In 2014, the introduction of deep learning and transfer learning in particular, ushered a new era for saliency prediction after several years of stagnating performance. Similarly, there has been only gradual progress in the recent 3-5 years, notwithstanding the significant amount of models proposed during that time (Figure \ref{fig:over-the-years}). From a machine learning point of view, the task of saliency prediction is conceptually interesting as it requires well-calibrated probabilistic predictions that are less critical in the much more common setting of highly deterministic classification problems.

In this work, we significantly improve spatial saliency modeling by studying how to achieve well-calibrated probabilistic predictions. Beyond proposing a new state-of-the-art model, we make a systematic analysis of the extent to which higher ImageNet performance leads to higher performance in the saliency domain. Specifically, we utilize a broad range of  models that have achieved state of the art on ImageNet as fixed feature backbones for the saliency prediction task, using a pointwise nonlinear read out following the DeepGaze II architecture and learning schedule as described in \cite{kummererUnderstandingLowHighLevel2017}. Additionally, we study the complementarity between these models and leverage it by conducting an ensemble learning approach which ends up yielding a new state of the art, closing the gap between models and inter-observer consistency in all metrics.

To gain additional insights into the differences between the backbones, we study the confidence calibration of the models based on them.
Confidence calibration is especially relevant when applying models in out-of-domain contexts where we would expect a good model to realize the domain shift and decrease its confidence accordingly \cite{ovadiaCanYouTrust2019}.
Many established confidence calibration measures \cite{guoCalibrationModernNeural2017} are not applicable in situations of very high stochasticity such as fixation prediction, therefore we propose a new method for testing confidence calibration which can be applied on datasets with high entropy.
Instead of being well calibrated or conservatively underconfident, we find that most individual models are highly overconfident on out-of-domain data, while our ensemble models show much better confidence calibration, which makes them more trustworthy on unseen datasets.


\section{Related Work}

Classic models have relied on hand engineered features to tackle saliency prediction \cite{itti1998model, Torralba, AIM, Kienzle, SUN}. Saliency prediction has since then moved on to deep learning models, the first of which was eDN \cite{eDN}. However one major hurdle when tackling saliency prediction with deep learning models is the small size of available data, stemming from the fact that collecting fixation data is both time-consuming and expensive. On top of that, the true saliency of an image is liable to shift when transformations are applied on it, severely limiting potential augmentations \cite{gazeGAN}. The first work that applied transfer learning to the saliency domain was DeepGaze I \cite{deepgazei} which has since then evolved to DeepGaze II that was built on VGG19 \cite{kummererUnderstandingLowHighLevel2017}. After DeepGaze I virtually every high-performing saliency model used transfer learning, usually based on ImageNet. Among the works that have focused on a principled transfer learning scheme for saliency prediction in the past was \cite{huang2015salicon}, which trained a saliency model on deep features from three CNNs (AlexNet, GoogleNet, and VGG16), combining low and high level pre-trained features, with a support vector machine on top and DeepFeat \cite{deepfeat} where the authors used a \textit{fixed} architecture on top of three pretrained CNN's features (ResNet, VGG, GoogleNet) to predict saliency. The EML-NET\cite{sotaEML} model introduced a scalable method to combine multiple deep convolutional networks of any complexity as encoders for features relevant to visual saliency.

Other models engineered complex deep architectures or build upon existing ones that have shown merit in other tasks, but all of them used transfer learning by pretraining their architectures with larger datasets as a starting point. SalGAN \cite{pan2017salgan} and GazeGAN \cite{gazeGAN} both used adversarial losses to train their saliency prediction models, which in the first case consist of an encoder-decoder architecture while the second one built on a U-net structure. The MSI-NET\cite{MSI-NET} tackled the task by integrating global scene information in its encoder-decoder architecture. UNISAL unified the image and video modalities of saliency to harness the entirety of saliency prediction datasets \cite{unisal}. 

Arguably, DeepFeat \cite{deepfeat} and EML-NET \cite{sotaEML} are the two works closest related to our own so contrasting with them makes it easier to highlight the finer shades of contribution in our work. DeepFeat uses a fixed linear readout on top of the pretrained features, whereas we fine tune a readout network that consists of 1$\times$1 convolutions following the DeepGaze II paradigm \cite{kummererUnderstandingLowHighLevel2017}. Features extracted by multilayer convolutional networks don't have a well defined scale due to many possible transformations, deeming the usage of a rigid linear readout too constrained for this type of input. In contrast, a readout network of 1$\times$1 convolutions is able to learn nonlinear transformations adjusting the scale of the input features and leverage interactions between those features. The small kernel size means that the network is unable to learn new spatial features but rather combines the ones given as input, making it an ideal tool for comparing the feature predictivity between different backbones for any given task. Aside from this major difference, we also conduct a series of studies that reveal how different models perform differently and combine their fixation densities to leverage their complementarity.

EML-NET aimed at maximal prediction performance, but we aim at understanding how much relevant information about fixation placement is encoded in deep features. To that end, we compare not only two but a large number of relevant ImageNet trained models. EML-NET is training each CNN model at the encoder stage while we keep ours fixed, which not only is less costly but also a much stronger scientific tool for studying the generalizeability of ImageNet trained features. Added to that, EML-NET combines these models at the encoder stage for a more broad prior knowledge, while in our case we study each model separately, delineating their individual contribution and later combine their predicted densities instead. 

Finally, compared to either of these works, we use a much broader array of state of the art ImageNet CNNs as backbones to our architectures and we train an agglomeration of each model configuration, accounting for the uncertainty in our metrics.

\section{Methods}

\subsection{Model and Training Pipeline}

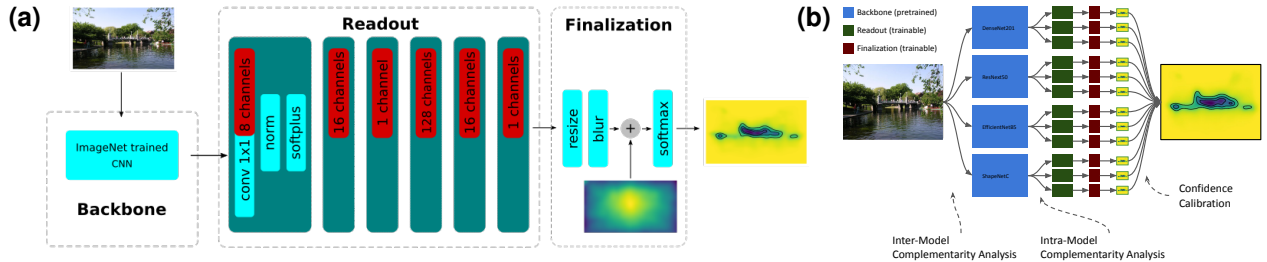
\begin{figure*}[htbp]
\begin{center}

 \begin{tikzpicture}
            \coordinate (hsep) at (0.2, 0.0);
            \coordinate (vsep) at (5, -5);
            \coordinate (labelsep) at (-0.5, -0);
            \tikzset{label/.style={font=\sffamily\bfseries}};
            \tikzset{anchor=north west};

            \node (model) at (0, 0) {\includegraphics[width=0.58\linewidth]{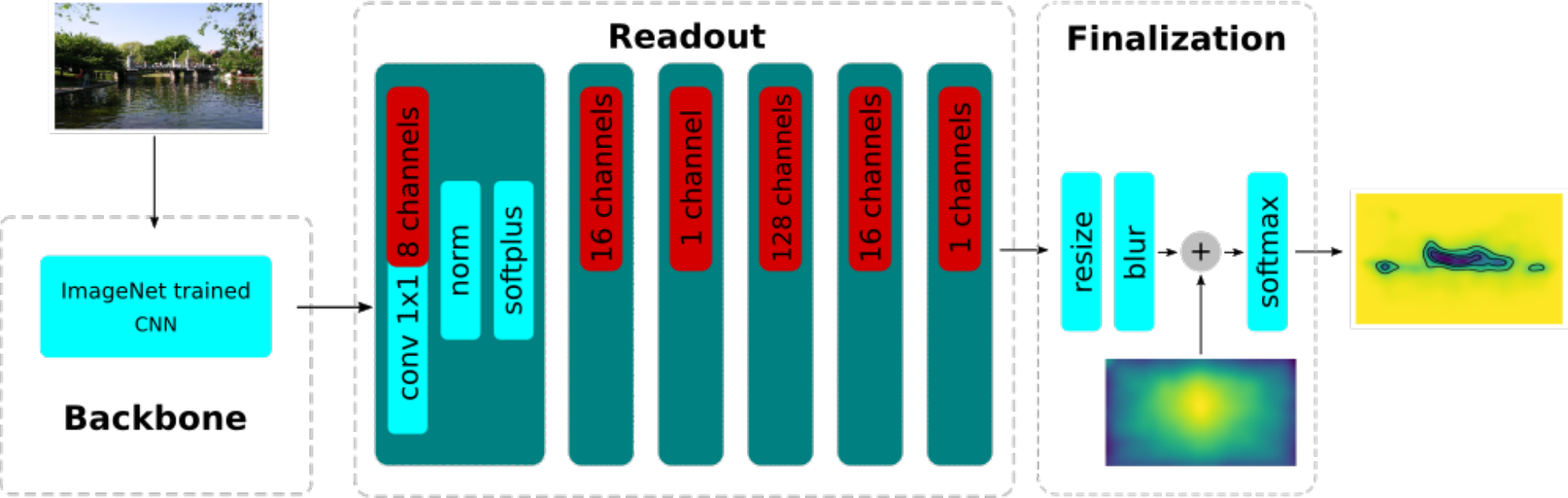}};
            \node (mixture) at ($ (model.north east) + (hsep) $) {\includegraphics[width=0.32\linewidth]{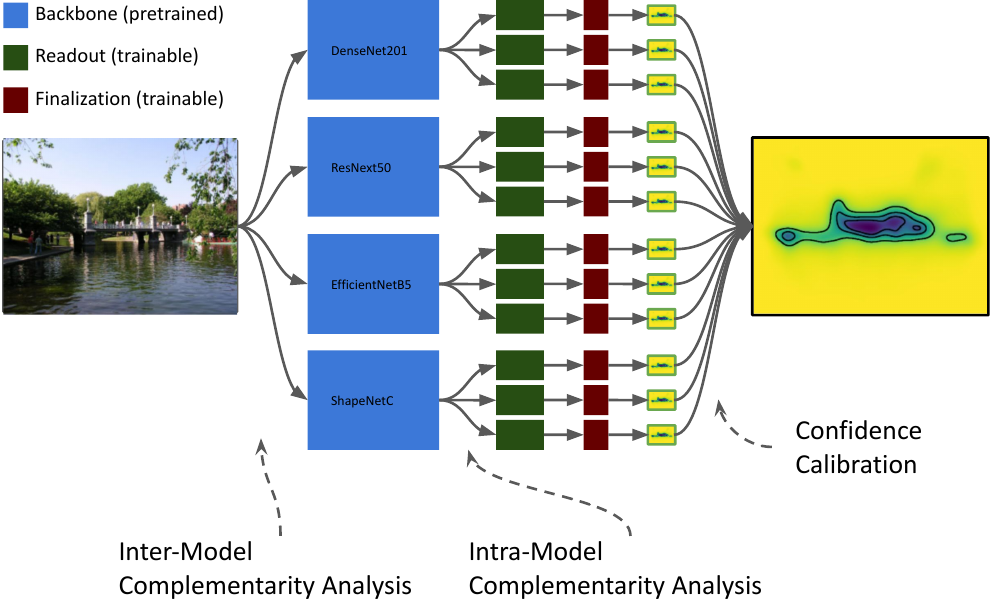}};
            
            \node[label] at ($ (model.north west) + (labelsep) $) {(a)};
            \node[label] at ($ (mixture.north west) + (labelsep) + (-0.1, 0) $) {(b)};
        \end{tikzpicture}

\caption{A diagram of our adapted DeepGaze II architecture as was used in all our experiments, as well as our best performing variant, DeepGaze IIE. (a) testing backbones: we collect some layers from CNNs pretrained on ImageNet without any additional training. We apply a readout network on these layers that consists of blocks of 1$\times$1 convolutions, a layernorm and a softplus function. 
Afterwards, a blur and a center bias prior are applied before a softmax that gives us the final probability density of fixations. (b) The ensemble model DeepGaze IIE: we combine some of the state of the art ImageNet backbones, leveraging inter- and intra- model complementarity which is analyzed in section 4.2. Confidence Calibration is used as an analytical tool to better understand why these models perform best.}
\label{fig:architecture}
\end{center}
\end{figure*}


The overall pipeline is visualized in Figure~\ref{fig:architecture}, where the final model is derived from the combination of multiple backbones after a series of principled analyses steps. An image is first processed with a backbone CNN to extract deep activations, which are subsequently processed in a readout network of $1
\times1$ convolutions.
The single output channel of the readout network is blurred, combined with a centerbias and fed through a softmax to yield a two-dimensional fixation distribution  (Figure~\ref{fig:architecture}a).
Essentially, this is an adaptation of the architecture of DeepGaze II \cite{kummererUnderstandingLowHighLevel2017} with a deeper readout network, layer norm and softplus instead of ReLUs as activation function, and, most notably, with different backbones instead of the original VGG19 network.
The readout network, along with the blur size and the centerbias weight are the only parts of the pipeline that undergo training.
The feature extractor's weights are kept fixed during this process.
Since our model predicts a fixation density, we have direct access to the likelihood of fixations and therefore we optimize our model for maximum likelihood. We first pretrain our model on the SALICON dataset \cite{SALICON} and then finetune it on the MIT1003 dataset \cite{MIT1003}. SALICON includes 10,000 images whose ground truth was collected using mouse traces as indicated by observers rather than a gaze detector. Albeit this seems to sacrifice precision, SALICON makes a good starting point as it has been proven to be very useful for pretraining saliency models. MIT1003 is composed of 1003 natural images tested on 15 subjects (with a presentation time of 3 seconds). The dataset contains images of various dimensions which we resized to either 1024$\times$786 or 768$\times$1024. Images were downsampled by a factor of 1.5 for SALICON and 2.0 for MIT1003/MIT300. We use a learning rate scheduler starting with an initial learning rate of 0.001 which then decays by a factor of 10 every set number of epochs.

We evaluate each configuration of our model following a 10-fold cross validation scheme on the MIT1003 dataset. In simple terms, given an MIT1003 image there is exactly one out of the ten models from this procedure that did neither see this image during training nor for validation in hyperparameter tuning, so that its predicted density is suitable for evaluation. Thus all reported metrics reflect test performance.

\subsection{Metrics}
As our main guide during our experiments, we used the Information Gain metric \cite{IGain}, which is effectively the difference in average log-likelihood of the model and a baseline model. Therefore, the metric measures the extent by which a model's knowledge has surpassed that of the baseline model. Since it's known that human fixations tend to accumulate towards the center of an image, we use an image-invariant center bias as baseline model.

For a model which predicts a fixation density $p(x \mid I)$ over possible fixation locations $x$ given an image $I$, the information gain is computed as
\[IG(\mathrm{model}) = \frac1N \sum_i^N \log_2 p_\mathrm{model}(x_i \mid I_i) - \log_2 p_\mathrm{baseline}(x_i),\]
where $x_i$ is the $i$th fixation of the dataset, taking place in image $I_i$.

We consider Information Gain the most principled metric \cite{IGain} and thus mostly rely on it, but we evaluate on other commonly used saliency metrics later on. These include AUC, shuffled AUC, KL divergence, Correlation Coefficient, and Normalized Scanpath Saliency \cite{bylinskiiWhatDifferentEvaluation2018}. Saliency metrics are known to be quite inconsistent when evaluated on the same saliency map \cite{bylinskiiWhatDifferentEvaluation2018}. However, it has recently been shown that this problem can be mitigated for probabilistic models by evaluating each metric on the saliency maps which has highest expected performance under the fixation density predicted by the model \cite{kummererSaliencyBenchmarkingMade2018}.
\begin{figure*}[t]
    \begin{center}
        \begin{tikzpicture}
            \coordinate (hsep) at (-0.4, 0);
            \coordinate (vsep) at (0, -5);
            \coordinate (labelsep) at (-0.1, -0.0);
            \tikzset{label/.style={font=\sffamily\bfseries}};
            \tikzset{anchor=north west};

            \node (calibmit) at (0, 0) {\includegraphics[width=0.25\linewidth]{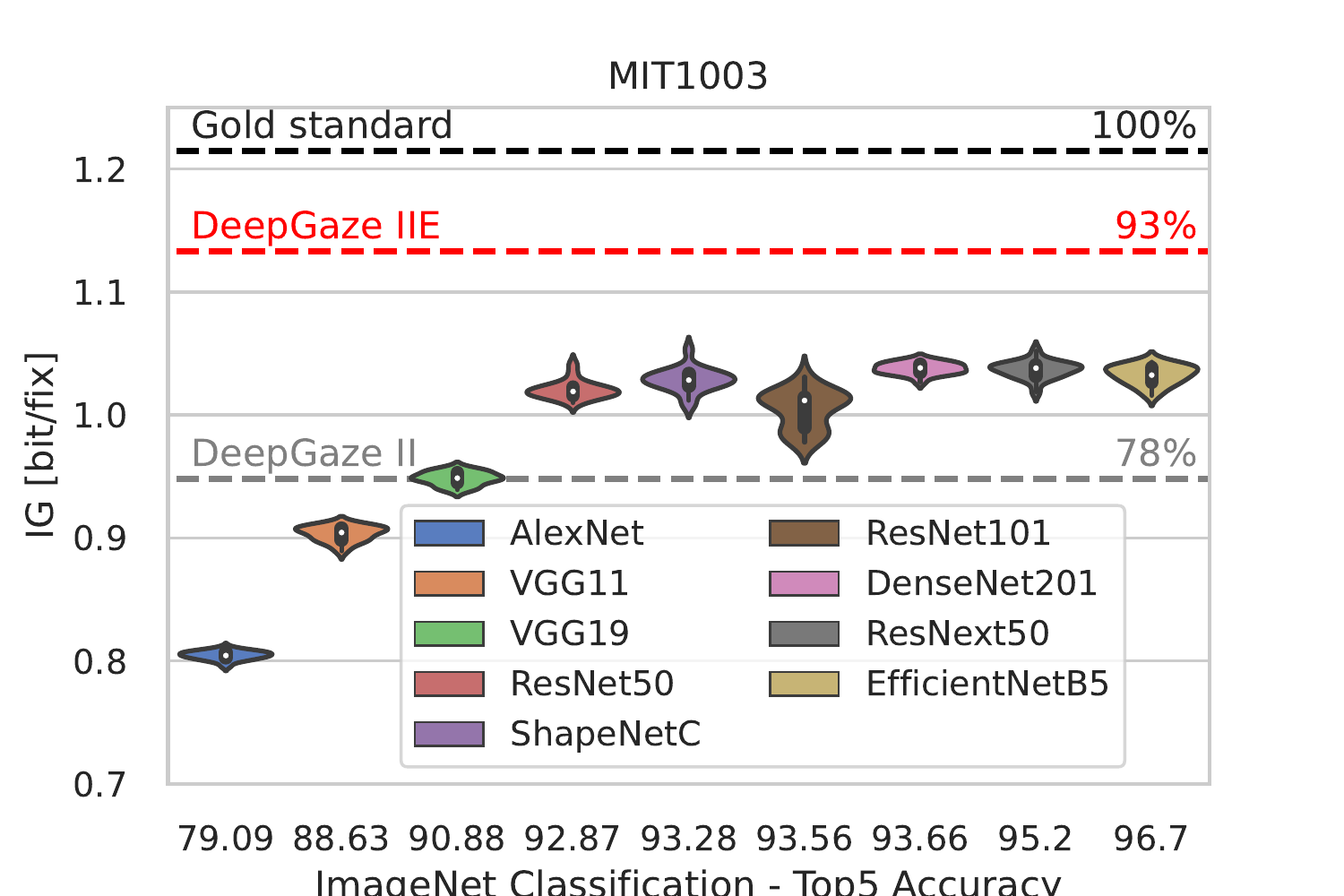}};
            \node (calibsalicon) at ($ (calibmit.north east) + (hsep) $) {\includegraphics[width=0.25\linewidth]{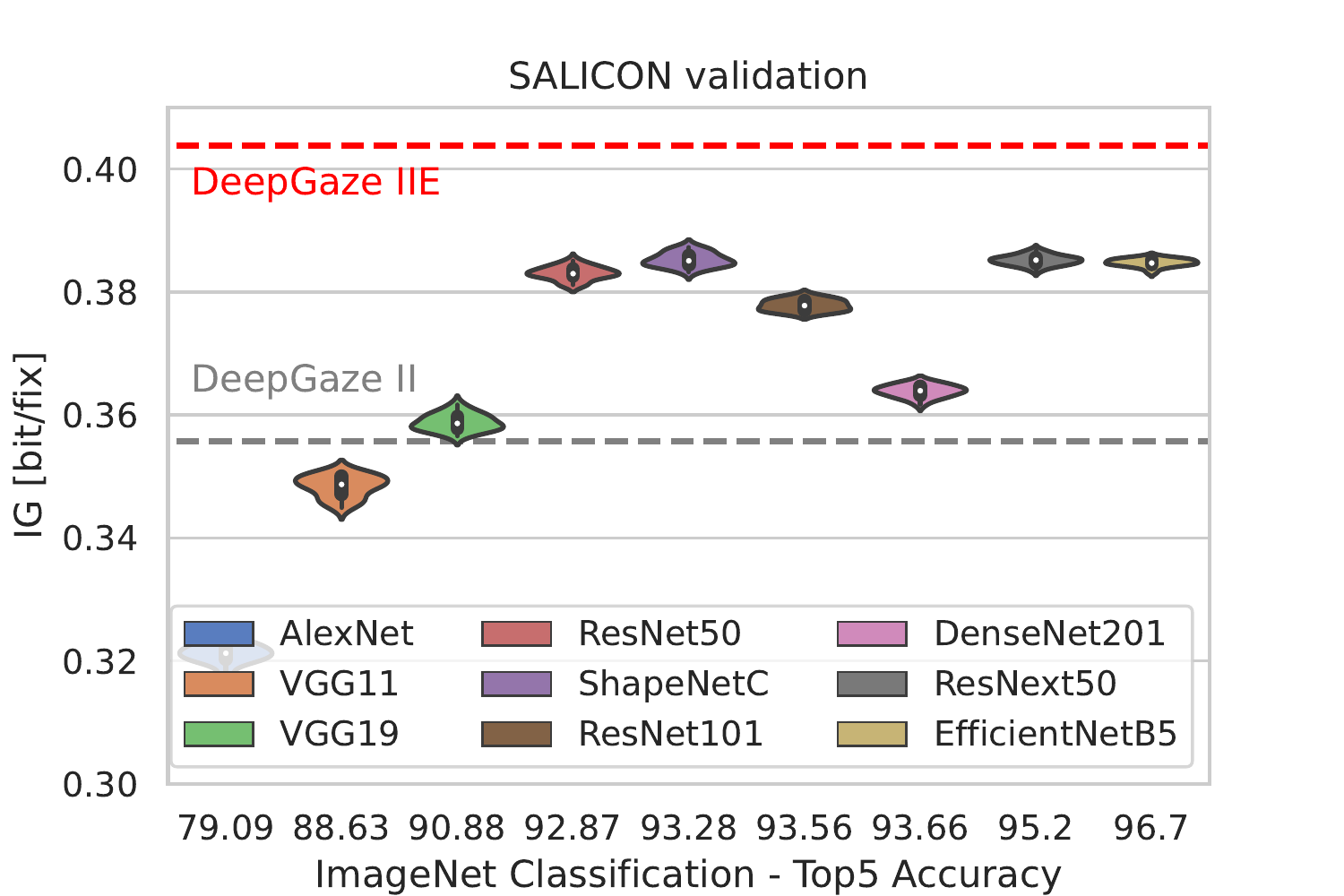}};
            \node (calibpascal) at ($ (calibsalicon.north east) + (hsep) $) {\includegraphics[width=0.25\linewidth]{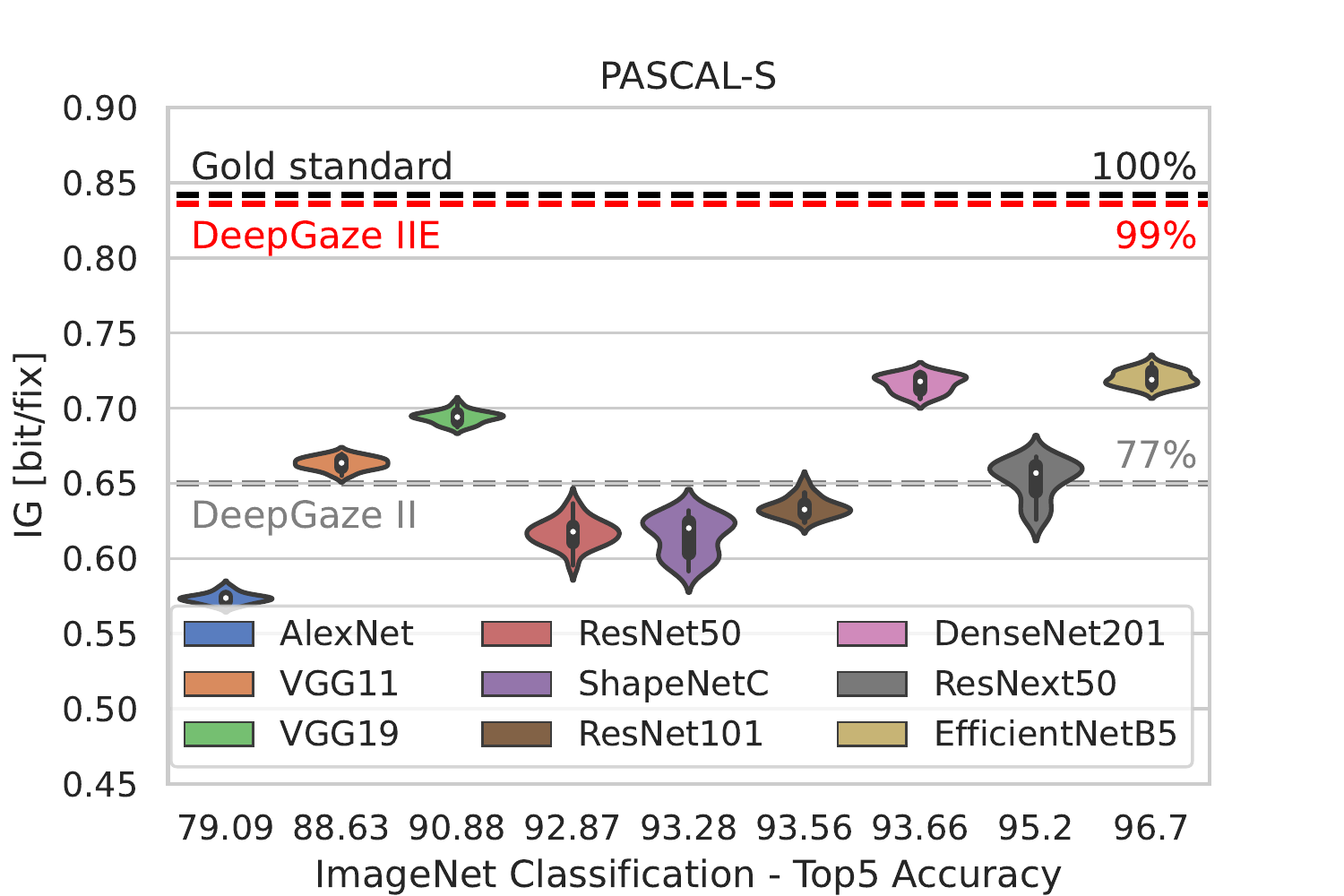}};
            \node (calibtoronto) at ($ (calibpascal.north east) + (hsep) $) {\includegraphics[width=0.25\linewidth]{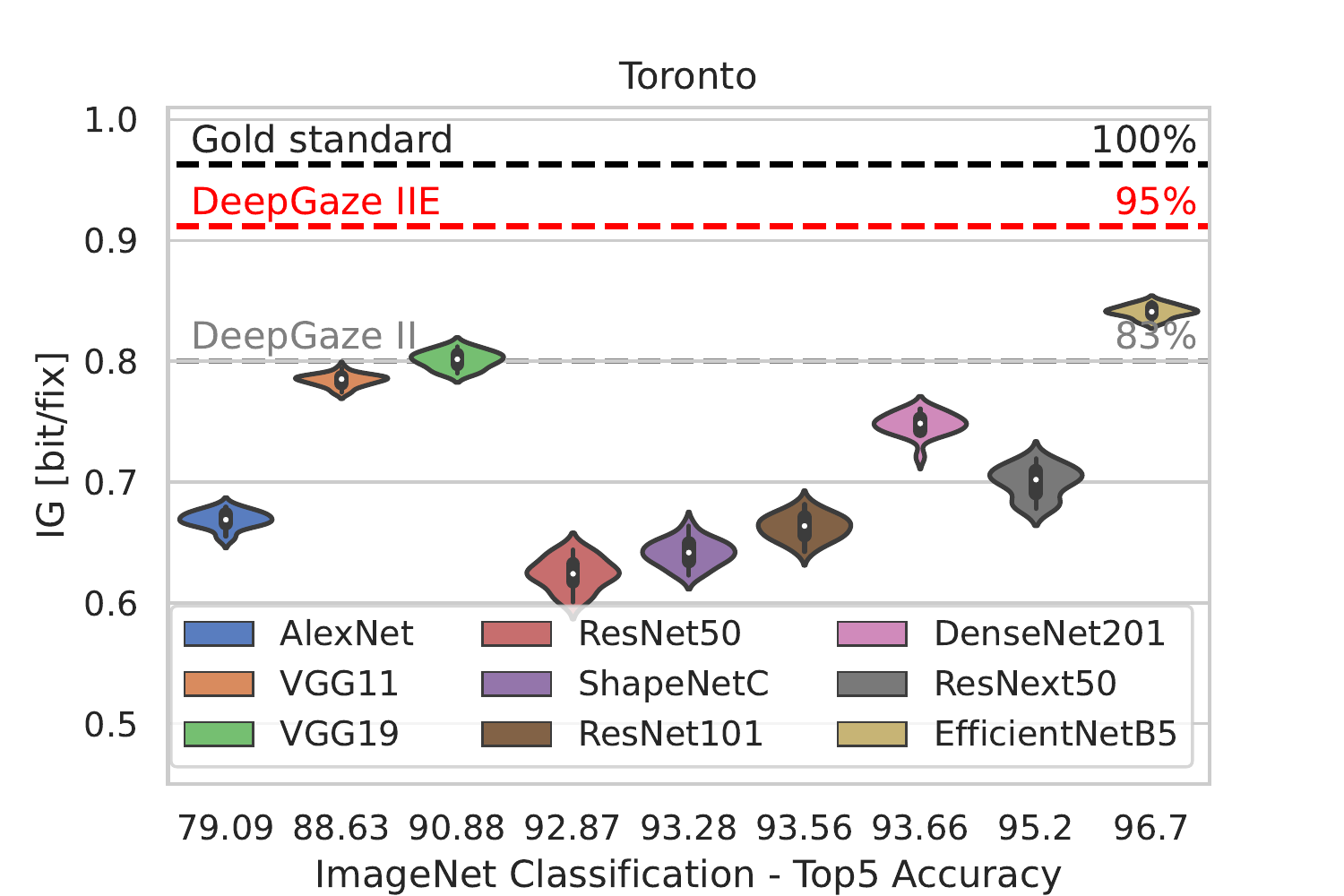}};
            
            \node[label] at ($ (calibmit.north west) + (labelsep) $) {(a)};
            \node[label] at ($ (calibsalicon.north west) + (labelsep) $) {(b)};
            \node[label] at ($ (calibpascal.north west) + (labelsep) $) {(c)};
            \node[label] at ($ (calibtoronto.north west) + (labelsep) $) {(d)};
        \end{tikzpicture}
    \end{center}
\begin{center}

\caption{Saliency prediction performance compared to ImageNet accuracy of model backbones. (a) Every violin plot is a representation of the performance distribution of 20 instances that share the same configuration and differ only in the initialization seed of their training. 
The red dashed line (DeepGaze IIE) indicates the performance of the best model we present in this paper (93\%), which averages multiple instances of models with different backbones.
The black dashed line (Gold standard = 100\%) shows an estimate of the achievable performance by the means of a nonparametric Gaussian KDE model of the fixations. The gray dashed line indicates the performance of the existing DeepGaze II model (78\%).
(b): Same on the SALICON validation dataset (using the models after the pretraining phase on SALICON). (c) and (d) We evaluate all models of (a) without retraining on the PASCAL-S dataset and the Toronto dataset. 
}


\label{fig:imnet-saliency comparison}
\end{center}
\end{figure*}
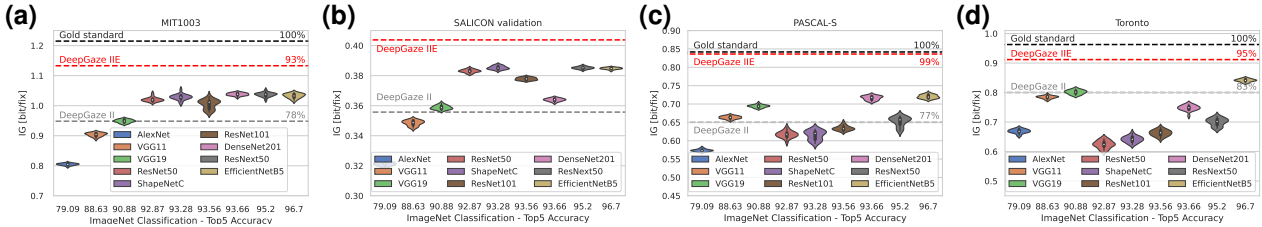

\subsection{Testing Confidence Calibration}
\label{sec:methods_calibration}

One key feature of saliency models is that they predict probabilistic fixation distributions rather than deterministic classes.
This means that our models predict not only qualitatively which regions they expect to be fixated, but also quantitatively how much more often they expect a certain salient region to be fixated than any other given region.
By comparing this to the actual numbers of fixations in the low-density and high-density regions, we can check how well calibrated the model confidence is---i.e. whether it makes overconfident or underconfident predictions.
Confidence calibration has been tested before for deep neural networks on classification tasks like ImageNet \cite{guoCalibrationModernNeural2017},
and it is known that deep neural networks have a tendency to make overconfident predictions on IID data and even more so on OOD data \cite{hendrycks*AugMixSimpleData2019, ovadiaCanYouTrust2019, heinWhyReLUNetworks2019}.
Ensembles trained on different augmentation techniques can mitigate this overconfidence to a certain degree \cite{sticklandDiverseEnsemblesImprove2020, ovadiaCanYouTrust2019}. 

Confidence calibration of classification models is usually tested using the Expected Calibration error, which compares a model's accuracy to its average confidence.
If a model is perfectly calibrated, its average confidence matches its accuracy.
Fixation prediction can be seen as a high-dimensional classification task where each image pixel constitutes a different class.
However, on ImageNet or similar classification tasks, usually only one or very few classes contain most of the probability mass, whereas, in fixation prediction, stochasticity is much higher such that even the most salient pixels have relatively low probability and differences between all pixels are relatively subtle. 
While this stochasticity makes confidence calibration even more important, accuracy will always be very low and is therefore  impractical to be used for an empirical test of calibration.
Instead, here we propose an approach which is more suited towards settings with high entropy.
First, we sort the pixels of a predicted fixation density by probability and then split them into multiple bins, each of identical probability mass.
For example, in Figure \ref{fig:teaser}, the model prediction is split by contour lines into four areas of decreasing size (yellow via green to blue), each of which accumulates 25\% of predicted fixation probability.
After segmenting the predicted fixation density, we count the empirically measured fixations in each area.
If the model is well calibrated, each area should receive the same number of fixations.
If the model is overconfident, it would assign a high probability to a region that would receive less than expected fixations, while other regions would receive more than expected fixations.
By averaging the number of fixations for each probability quantile, i.e., area, over the full dataset, we can summarize the confidence calibration in a histogram.

\section{Experiments and Results}

\subsection{Transferring ImageNet Features to Saliency in a Principled Way}

Our set-up is outlined as follows: first we obtain an architecture trained on ImageNet classification and train a readout network that takes as input a certain number of deep layers whose total number of channels are approximately 2048 (see below for details on the layer selection strategy). These layers are either convolutional or activation layers (ReLU). We use the following networks as backbones for our readout network: AlexNet \cite{AlexNet}, VGG11 and VGG19 \cite{VGG}, ResNet50 and ResNet101 \cite{resnet}, ShapeNet \cite{ShapeNet}, EfficientNet-B5 \cite{efficentNet} and DenseNet \cite{densenet}. Note that regarding ShapeNet, there are 3 configurations regarding how the model was trained, and we choose the one trained on ImageNet and Stylized-ImageNet, then fine tuned on ImageNet. We will be referring to it as ShapeNet-C.

\subsubsection{Selecting Layers From the Backbone}



For each network, we conduct two sets of experiments: first we deduce which extracted layers are leading to the best performance (\textit{layer search} stage), then repeat multiple initializations of the exact same configuration to gain a robust metric of final performance (\textit{instance search} stage). 
Our preliminary results showed that fluctuations appear even between different instances of the same layer configuration and are of the same magnitude as fluctuations between the top 5 performing layer configurations, indicating that an extensive layer search has marginal value.
Thus, we test 10 possible configurations during layer search followed by a training of 20 instances from the top configuration. Given that there appear to be notable fluctuations even between different instances of the same pipeline, evaluating information gain across 20 instances gives us a more robust picture of a model's true performance and an estimate of its epistemic uncertainty.

In general, we find that using approximately 3-4 layers from the ultimate and penultimate layer spaces is ideal while using a single layer consistently results in highly suboptimal performance.

\subsubsection{ImageNet accuracy as an indicator of saliency prediction performance}
In Figure \ref{fig:imnet-saliency comparison}a we show the prediction performances of each backbone on the MIT1003 dataset. For each backbone we show the performance distribution of the 20 trained instances. Our results show that ImageNet performance transfers linearly to saliency up until it reaches a plateau. 
Specifically, we see a big leap in saliency performance starting from AlexNet and leading up to ResNet-50 which then slows down until it peaks at DenseNet-201, dropping off afterwards.
This trend is also visible in all other commonly used saliency metrics (Supplementary Material, Table 1)

\subsection{Investigating Model Complementarity}

When two distinct models perform almost as well on a dataset, there are two potential assumptions: One, that they are learning the exact same pieces of information and thus achieve similar performance, likely one of them doing it in a slightly better way. Two, the models are doing equally well on the whole dataset but might be achieving that by encoding different and potentially complementary pieces of information. 




\subsubsection{Mixtures of Fixation Densities}

In additional experiments we found substantial variances in per-image performance both for models with different backbones and for model instances using the same backbone but different random seeds (see Supplementary Material).
This suggests that not only the different backbones but also the different instances of the models using the same backbones in our experiments encode different information.
This finding motivated us to leverage the apparent complementarity of the information our models encode in terms of  \textit{inter-model complementarity} (different backbones) and \textit{intra-model complementarity} (different instances within the same model).
We average the predicted fixation densities in a pairwise manner across some of our probabilistic models, varying the weights of each predicted density. After conducting this experiment in several pairwise combinations, we find that we consistently get an improvement in performance that peaks when the two models have equal weights (Supplementary Material, Figure 3). 

We sought to leverage inter-model complementarity by combining all of our top performing models in a pairwise manner, then triple-wise and finally a quadruple-wise mixture of ShapeNet-C, EfficientNet-B5, ResNext-50, DenseNet-201 (weights being equal in all cases). Model performances consistently improve when adding models until the quadruple mixture achieves top performance (Table \ref{tab:inter-complementarity}). Adding ResNet-50 for a total of 5 model mixture decreases performance and therefore we stop at four backbones.


\begin{figure*}
    \begin{center}
        \begin{tikzpicture}
            \coordinate (hsep) at (5.8, 0);
            \coordinate (vsep) at (0, -5);
            \coordinate (labelsep) at (-0.3, -0.0);
            \tikzset{label/.style={font=\sffamily\bfseries}};
            \tikzset{anchor=north west};

            \node (calibmit) at (0, 0) {\includegraphics[width=0.30\linewidth]{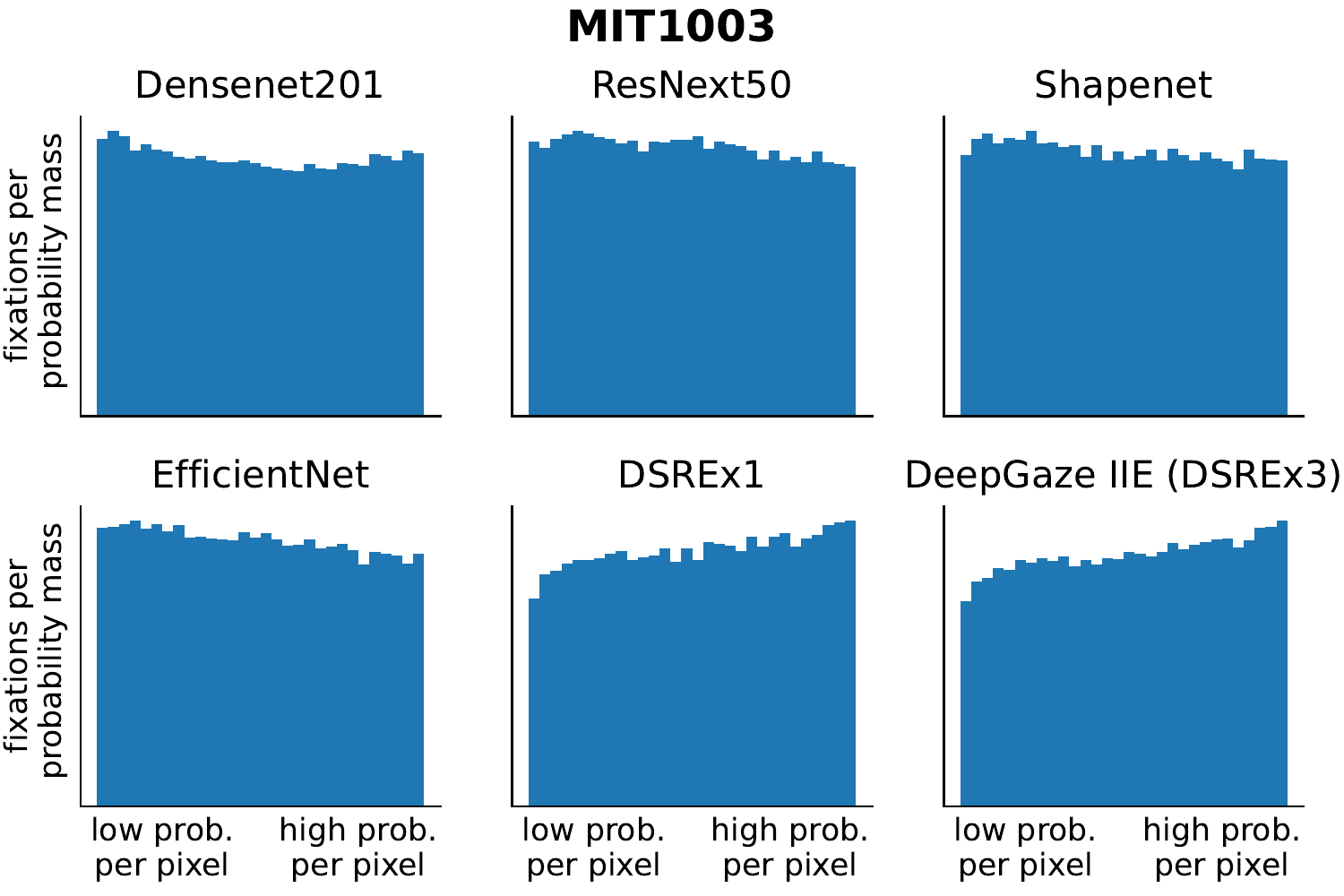}};
            \node (calibpascal) at ($ (calibmit.north west) + (hsep) $) {\includegraphics[width=0.30\linewidth]{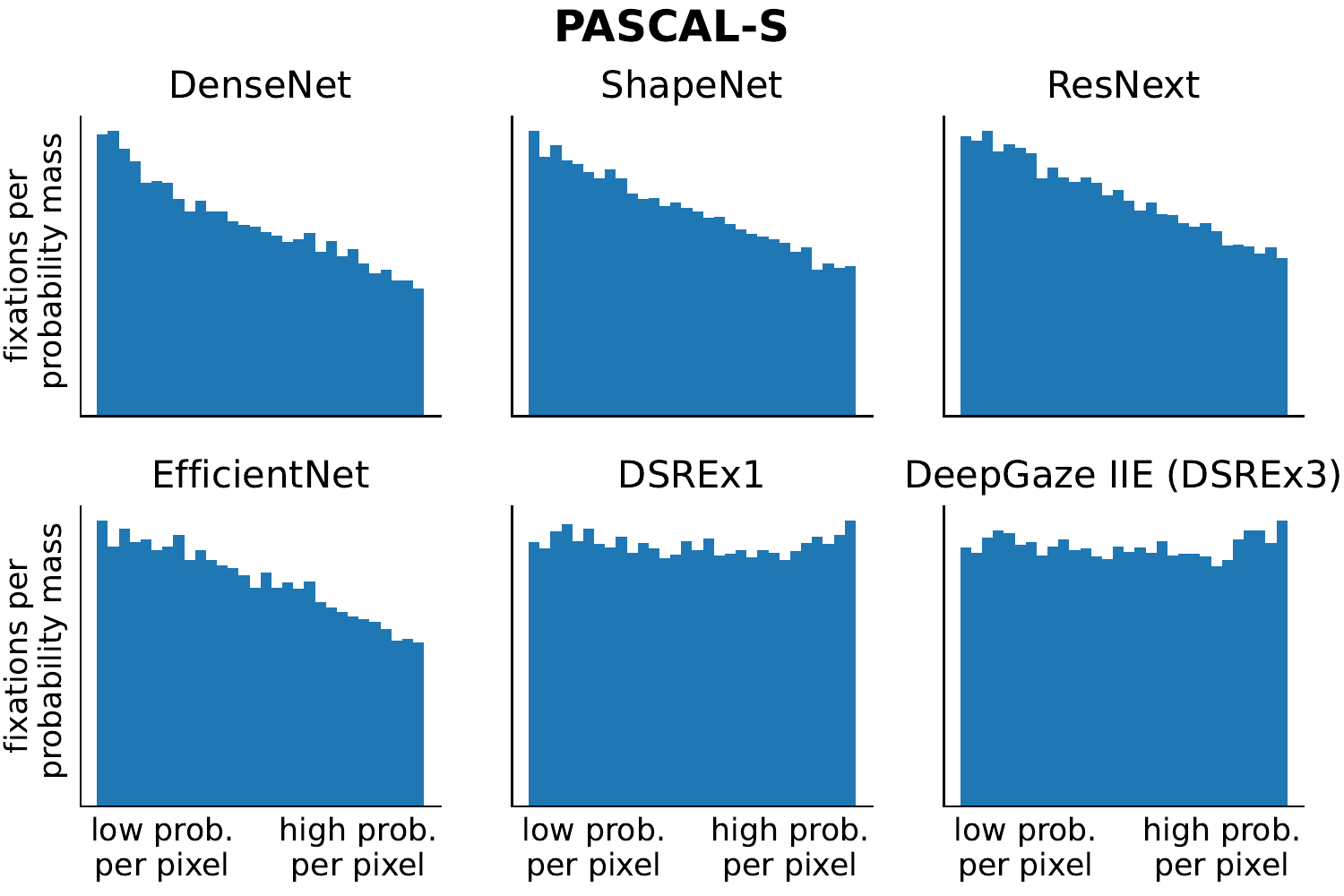}};
            \node (calibtoronto) at ($ (calibpascal.north west) + (hsep) $) {\includegraphics[width=0.30\linewidth]{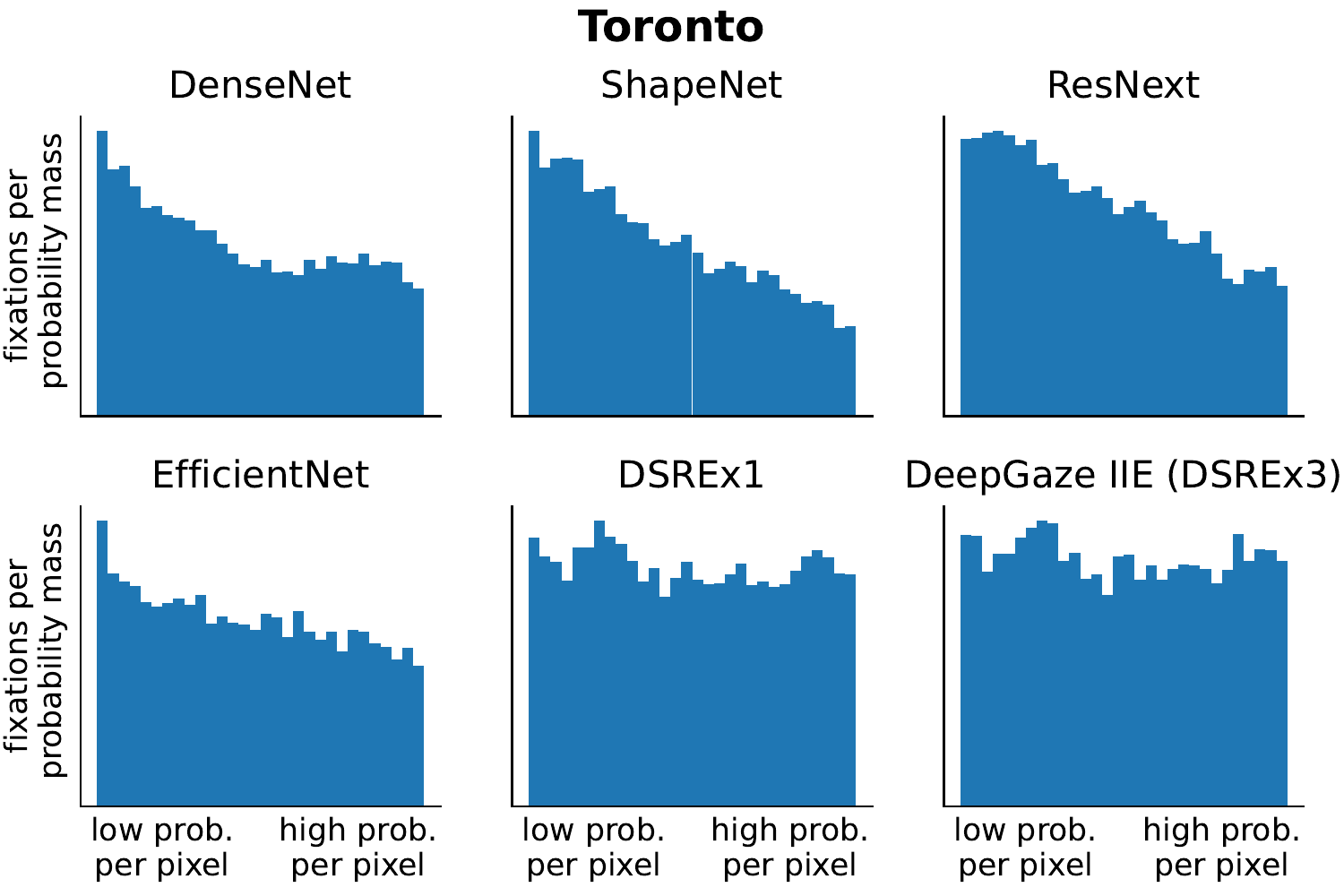}};
            
            \node[label] at ($ (calibmit.north west) + (labelsep) $) {(a)};
            \node[label] at ($ (calibpascal.north west) + (labelsep) $) {(b)};
            \node[label] at ($ (calibtoronto.north west) + (labelsep) $) {(c)};
        \end{tikzpicture}
    \end{center}
    \caption{Confidence calibration on different datasets (a: MIT1003, b: PASCAL-S, c: Toronto) for different models (individual histograms). We split predicted fixation densities into multiple quantiles of identical probability mass but sorted by increasing probablity per pixel and quantify the number of actual fixations per predicted probability to assess whether models are overconfident (bar heights decreasing from left to right), underconfident (bar heights increasing from left to right) or well calibrated (uniform histogram). On MIT1003, the dataset used for training, models with individual backbones are quite well calibrated and ensemble models (DSREx1 and DeepGaze IIE=DSREx3) are slightly underconfident. In the generalization setting on PASCAL-S and Toronto, individual models are strongly overconfident while the ensemble models are close to perfectly calibrated.}
    \label{fig:calibration}
\end{figure*}
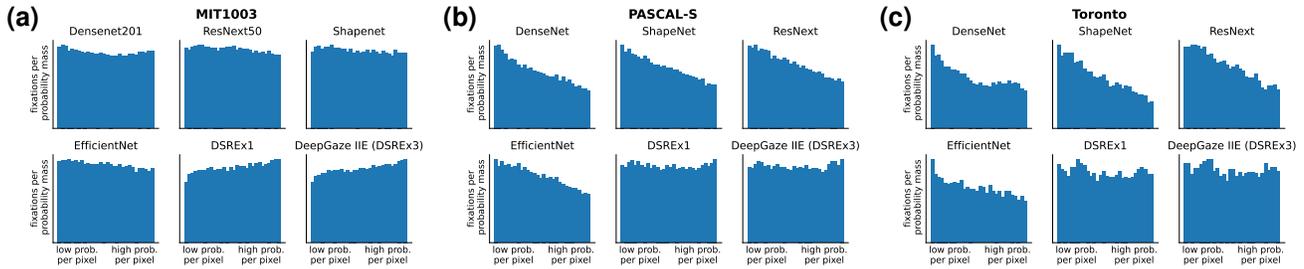

\begin{table*}[htbp]

    \begin{center}
    
    \caption{Leveraging \textit{inter}-model complementarity: We mixed our top performing models, starting with pairwise mixtures and leading up to a mixture of four. Note that for illustration purposes, darker shades of red represent more components for the corresponding mixture model.}
    \label{tab:inter-complementarity}
    \begin{tabular*}{\textwidth}{l@{\extracolsep{\fill}}cccc}
    \toprule
    Backbones & None & DenseNet-201 & EfficientNet-B5 & DenseNet-201, EfficientNet-B5  \\
    \midrule
    None & & \textcolor{pale}{1.0377} & \textcolor{pale}{1.0326} & \textcolor{light}{1.1077} \\
    ResNext-50  & \textcolor{pale}{1.0368} & \textcolor{light}{1.1075}  & \textcolor{light}{1.1052} & \textcolor{dark}{1.1256}\\
    ShapeNet-C  & \textcolor{pale}{1.0278} & \textcolor{light}{1.1025} & \textcolor{light}{1.0986} & \textcolor{dark}{1.1213} \\
    ResNext-50, ShapeNet-C & \textcolor{light}{1.0904} & \textcolor{dark}{1.1165} & \textcolor{dark}{1.1143} & \textbf{\textcolor{darkest}{1.1285}}\\
    \bottomrule
    \end{tabular*}
    \end{center}

\end{table*}

\begin{table*}[htbp]
    \centering
    \caption{Leveraging \textit{intra}-model complementarity: We split the four model mixture (DSRE) into more instances per model and evaluate for each number of instances.}
        \begin{tabular*}{\textwidth}{l@{\extracolsep{\fill}}ccccc}
        \toprule
        Number of instances & 1 & 2 & 3 & 4 & 5  \\
        \midrule
        DSRE  & 1.1285  & 1.13193 &  \textbf{1.13294} & 1.13285 &  1.13287 \\
        \bottomrule
        \hline
        \end{tabular*}
    \label{tab:intra-model-complementarity}
\end{table*}

As even within the same backbone, there is a significant variance in per sample performance, we exploit not only inter-model complementarity but also intra-model complementarity. To do that we keep the 4 backbones we found to be best and for each of them average several instances, effectively averaging for 4 models $\times$ 2 instances, then 4 models $\times$ 3 instances etc leading up to 5 instances per model for a total mixture of 20 instances. The split does not change each model's impact on the total average, but rather makes it so each model has a more educated decision by averaging over a greater number of its instances. Leveraging intra-model complementarity, we achieve further boost in performance that saturated at 3 instances per model with a final information gain score of 1.1329 bit/fixation compared to 1.1285 bit/fixation for only one instance per model (Table \ref{tab:intra-model-complementarity}).
This best performing model DSREx3 will be called ``DeepGaze IIE'' in the following (``E'' for ``ensemble'').
In the Supplement, Figure 4, we visualize example predictions of the different models.

\begin{table*}[htbp]
    \centering
    \caption{Models scores on the MIT300 benchmark. Notably, some models are missing IG as they are not probabilistic and thus impossible to evaluate under this metric. DINet is not included in the public MIT300 leaderboard, therefore we show the scores reported in their paper.}
        \begin{tabular*}{\textwidth}{l@{\extracolsep{\fill}}ccccccc}
        \toprule
        Model &      IG $\uparrow$ &     AUC $\uparrow$ &    sAUC $\uparrow$ &     NSS $\uparrow$ &      CC $\uparrow$ &   KLDiv $\downarrow$ & SIM $\uparrow$ \\
        \midrule
        DeepGaze IIE (DSREx3)  &  \textbf{1.0715} & \textbf{0.8829} &  {0.7942} &  \textbf{2.5265}  &  \textbf{0.8242} &  \textbf{0.3474} & {0.6993}\\
        DSREx1 & 1.0679  & 0.8825  & 0.7938 & 2.5219 & 0.8234 & 0.3489 & 0.6987\\
        UNISAL+EML-Net+MSI-Net & 1.0607 & 0.8824 & \textbf{0.7948} & 2.5131 & 0.8239 & 0.3537 &  \textbf{0.7030} \\
        \midrule
        UNISAL \cite{unisal} & 0.9505 & 0.8772 &  	 	0.7840  & 2.3689 & 0.7851 & 0.4149 &  	0.6746\\
        EML-NET \cite{sotaEML} &   & 0.8762 &  0.7469 &  2.4876  &  0.7893 &  0.8439 & 0.6756 \\
        MSI-NET \cite{MSI-NET} &   0.9185  & 0.8738  &  0.7787 &  2.3053  & 0.7790 & 0.4232 & 0.6704 \\
        DeepGaze II \cite{kummererUnderstandingLowHighLevel2017} &  0.9247 & 0.8733 &  	0.7759 & 2.3371  & 0.7703 &  	0.4239 & 0.6636\\
        TranSalNet &  &  0.8730  & 0.7471  &  2.3758  &  0.7991  &  0.9019 &  0.6852 \\
        GazeGAN \cite{gazeGAN} &  &   0.8607   &  0.7316   &   2.2118   &   0.7579   &   1.3390  &   0.6491  \\
        DINet \cite{yangDilatedInceptionNetwork2019} &  & 0.86 & 0.71 & 2.33 & 0.79 & & \\
        
        \bottomrule
        \hline
        \end{tabular*}
    \label{tab:MIT300_eval}
\end{table*}


\subsubsection{Generalization Performance}

In Figure \ref{fig:imnet-saliency comparison}c and d, we show how well the models with different backbones generalize to the PASCAL-S dataset \cite{PASCALS} and the Toronto dataset \cite{AIM}.
It can be seen that not all backbones generalize similarly well.
While VGG, DenseNet and EfficientNet show good generalization performance on both datasets, ResNet, ShapeNet and ResNext show substantially worse performance.
The DeepGaze IIE ensemble model again shows a substantial performance boost compared to all individual models, with performance close to the gold standard performance (99\% on PASCAL-S and 95\% on Toronto).
Especially on PASCAL-S, the performance gain relative to the best backbone (EfficientNet) is nearly as good as the performance difference between the best and the worst backbone.
In Figure \ref{fig:imnet-saliency comparison}b, we also show how well the models  with different backbones perform on the SALICON validation set (using model weights from pretraining on SALICON).
Here, again a very similar pattern can be observed.
Since SALICON is a much larger dataset than MIT1003, this provides additional evidence that DeepGaze IIE is not simply solving an overfitting problem but leverages different information from different backbones.

Finally, we also test our ensemble models on the held-out MIT300 dataset of the MIT/Tuebingen Saliency Benchmark \cite{BenchmarkMIT300, mit-tuebingen-saliency-benchmark}.
Our pairwise combination of models is already enough to beat the state of the art, while our final combination of four models with three instances each leads to an even higher leap on the state of the art.
The power of ensembling different models with different backbones is further demonstrated by a mixture of the three current top-performing models on MIT300 (UNISAL, EML-Net and MSI-Net), which also outperforms current state-of-the-art, but is still slightly outperformed by DeepGaze IIE.
The benchmark results are displayed at table \ref{tab:MIT300_eval}. 
In the Supplement, Table 2, we also report scores on the SALICON test set.

\subsubsection{Confidence Calibration}


In Figure \ref{fig:calibration}, we visualize confidence calibration for our models (see Section \ref{sec:methods_calibration} for details).
Uniform histograms indicate perfect confidence calibration while histograms skewed to the left indicate overconfident models: there are not as many fixations in high-saliency regions as expected by the model.
Histograms skewed to the right indicate underconfident models.
In Figure \ref{fig:calibration}a, we evaluate confidence calibration for models with four different backbones as well DSREx1 and DeepGaze IIE on the MIT1003 dataset.
Evidently, all individual backbones are fairly well calibrated (the histograms are close to uniform), with a slight bias towards overconfidence.
The ensemble models DSREx1 and DeepGaze IIE on the other hand are a bit underconfident.
When generalizing to the PASCAL-S and Toronto datasets \cite{AIM,PASCALS} (Figures \ref{fig:calibration}b and c), this effect changes: all individual models are now strongly overconfident, while the ensemble models are close to perfectly calibrated on both datasets. This suggests that individual models make different errors on new images, which are compensated by using an ensemble of models with different backbones. Interestingly, this doesn't hold when we exclusively average models with the same backbone. Apparently, the problem is not noise in the readout network, but overfitting to certain features of the backbone, which happen to be overly correlated with fixations on the MIT1003 dataset.
Since ensembling helps, features used by individual models likely differ substantially across backbones.






\section{Discussion}



Although the models trained for ImageNet classification contain features of high value to saliency prediction, features extracted from ImageNet classification have reached a point of diminishing returns where additional classification accuracy no longer clearly transfers to higher prediction performance in the saliency domain.
However, features from different backbones don't seem to be correlated with saliency in the same way.
This is suggested by the fact that models using different backbones generalize in very different ways to new datasets, and even more by the fact that ensemble models substantially outperform even the best individual models both within dataset and on new datasets.

In order to test how useful our models are in practical applications on unseen datasets, we test out-of-domain performance not only with respect to prediction performance, but also with respect to confidence calibration.
We find that our individual models tend to be substantially overconfident on out-of-domain data, while our ensemble models are slighlty underconfident on within-domain-data but close to perfectly calibrated on out-of-domain data, which makes them more applicable on unseen datasets.
The method which we propose for assessing confidence calibration can be easily applied in settings with a high number of classes and high stochasticity in the ground truth distribution.

With regards to saliency prediction, performance has somewhat stagnated in recent years thus making the observed leap even more significant, especially if we consider that our architectures are not overengineered to the task but rather are part of a principled pipeline that could potentially be applied in other domains. We attribute the success to four factors: First, our choice of readout network, which is less constrained than a linear readout allowing it to make nonlinear transformations of input features but more constrained than a typical CNN as it uses only 1$\times$1 kernels. This allows it to combine the spatial features without creating new ones making it an efficient tool for transfer learning and allowing interpretability in its results, since we don't finetune the backbones.
While finetuning the backbone in theory could result in even better performance, we found that by fine tuning the large parameter space we inevitably overfit MIT1003 and consistently produces worse results.
The second factor is our utilization of multiple instances of each model. We argue that this is good practice as it models the uncertainty in these models which in some cases such as ResNet101 is much higher than one would expect. Third, we leverage multiple models and combine them in a principled way utilizing both the complementarity between architectures and between instances of the same architectures which we labeled inter- and intra-complementarity respectively. For saliency this sort of combination is really simple, not requiring an oracle network but rather a simple averaging process of the fixation densities. Fourth, we used information gain to guide our experiments and have highlighted how relative performance transfers reliably to other metrics and other datasets. It has been argued that information gain is ideal for principled studies due to its foundation in information theory and its independence of hyperparameters \cite{IGain}. 
In the future, maximally diverse backbones should be further explored to yield even better models. This could be done through correlation analysis or by combining ImageNet backbones such as the ones presented here with self-supervised ones, as well as backbones pre-trained on other tasks such as object detection.

Taken together we have shown that our principled ensemble learning approach yields a 15 percent point improvement over DeepGaze II, setting the new state of the art in saliency prediction on the MIT/Tuebingen Saliency Benchmark in all available metrics, a significant leap after 4 years of only gradual progress, highlighting the promise of our approach.

{\small
\bibliographystyle{ieee_fullname}
\bibliography{egbib}

\begin{thebibliography}{10}\itemsep=-1pt

\bibitem{CAT2000}
Ali Borji and Laurent Itti.
\newblock Cat2000: A large scale fixation dataset for boosting saliency
  research.
\newblock {\em CVPR 2015 workshop on "Future of Datasets"}, 2015.
\newblock arXiv preprint arXiv:1505.03581.

\bibitem{AIM}
Neil Bruce and John Tsotsos.
\newblock Attention based on information maximization.
\newblock {\em Journal of Vision}, 7(9):950--950, 2007.

\bibitem{bylinskiiWhatDifferentEvaluation2018}
Z. Bylinskii, T. Judd, A. Oliva, A. Torralba, and F. Durand.
\newblock What do different evaluation metrics tell us about saliency models?
\newblock {\em {IEEE} Transactions on Pattern Analysis and Machine
  Intelligence}, pages 1--1, 2018.

\bibitem{gazeGAN}
Zhaohui Che, Ali Borji, Guangtao Zhai, Xiongkuo Min, Guodong Guo, and Patrick
  Le~Callet.
\newblock How is gaze influenced by image transformations? dataset and model.
\newblock {\em IEEE Transactions on Image Processing}, 29:2287--2300, 2019.

\bibitem{ImageNet}
Jia Deng, Wei Dong, Richard Socher, Li-Jia Li, Kai Li, and Li Fei-Fei.
\newblock Imagenet: A large-scale hierarchical image database.
\newblock In {\em 2009 IEEE conference on computer vision and pattern
  recognition}, pages 248--255. Ieee, 2009.

\bibitem{DECAF}
Jeff Donahue, Yangqing Jia, Oriol Vinyals, Judy Hoffman, Ning Zhang, Eric
  Tzeng, and Trevor Darrell.
\newblock Decaf: A deep convolutional activation feature for generic visual
  recognition.
\newblock In {\em International conference on machine learning}, pages
  647--655, 2014.

\bibitem{unisal}
Richard Droste, Jianbo Jiao, and J~Alison Noble.
\newblock Unified image and video saliency modeling.
\newblock {\em arXiv preprint arXiv:2003.05477}, 2020.

\bibitem{ShapeNet}
Robert Geirhos, Patricia Rubisch, Claudio Michaelis, Matthias Bethge, Felix~A
  Wichmann, and Wieland Brendel.
\newblock Imagenet-trained cnns are biased towards texture; increasing shape
  bias improves accuracy and robustness.
\newblock {\em arXiv preprint arXiv:1811.12231}, 2018.

\bibitem{guoCalibrationModernNeural2017}
Chuan Guo, Geoff Pleiss, Yu Sun, and Kilian~Q. Weinberger.
\newblock On calibration of modern neural networks.
\newblock In {\em International Conference on Machine Learning}, pages
  1321--1330. {PMLR}, 2017.
\newblock {ISSN}: 2640-3498.

\bibitem{resnet}
Kaiming He, Xiangyu Zhang, Shaoqing Ren, and Jian Sun.
\newblock Deep residual learning for image recognition.
\newblock In {\em Proceedings of the IEEE conference on computer vision and
  pattern recognition}, pages 770--778, 2016.

\bibitem{heinWhyReLUNetworks2019}
Matthias Hein, Maksym Andriushchenko, and Julian Bitterwolf.
\newblock Why {ReLU} networks yield high-confidence predictions far away from
  the training data and how to mitigate the problem.
\newblock {\em {arXiv}:1812.05720 [cs, stat]}, 2019.

\bibitem{hendrycks*AugMixSimpleData2019}
Dan Hendrycks*, Norman Mu*, Ekin~Dogus Cubuk, Barret Zoph, Justin Gilmer, and
  Balaji Lakshminarayanan.
\newblock {AugMix}: A simple data processing method to improve robustness and
  uncertainty.
\newblock In {\em Proceedings of the International Conference on Learning
  Representations (ICLR)}, 2019.

\bibitem{densenet}
Gao Huang, Zhuang Liu, Laurens Van Der~Maaten, and Kilian~Q Weinberger.
\newblock Densely connected convolutional networks.
\newblock In {\em Proceedings of the IEEE conference on computer vision and
  pattern recognition}, pages 4700--4708, 2017.

\bibitem{huang2015salicon}
Xun Huang, Chengyao Shen, Xavier Boix, and Qi Zhao.
\newblock Salicon: Reducing the semantic gap in saliency prediction by adapting
  deep neural networks.
\newblock In {\em Proceedings of the IEEE International Conference on Computer
  Vision}, pages 262--270, 2015.

\bibitem{itti1998model}
Laurent Itti, Christof Koch, and Ernst Niebur.
\newblock A model of saliency-based visual attention for rapid scene analysis.
\newblock {\em IEEE Transactions on pattern analysis and machine intelligence},
  20(11):1254--1259, 1998.

\bibitem{sotaEML}
Sen Jia and Neil~D.B. Bruce.
\newblock Eml-net: An expandable multi-layer network for saliency prediction.
\newblock {\em Image and Vision Computing}, 95:103887, 2020.

\bibitem{SALICON}
M. Jiang, S. Huang, J. Duan, and Q. Zhao.
\newblock Salicon: Saliency in context.
\newblock In {\em 2015 IEEE Conference on Computer Vision and Pattern
  Recognition (CVPR)}, pages 1072--1080, June 2015.

\bibitem{BenchmarkMIT300}
Tilke Judd, Fr{\'e}do Durand, and Antonio Torralba.
\newblock A benchmark of computational models of saliency to predict human
  fixations.
\newblock In {\em MIT Technical Report}, 2012.

\bibitem{MIT1003}
Tilke Judd, Krista Ehinger, Fr{\'e}do Durand, and Antonio Torralba.
\newblock Learning to predict where humans look.
\newblock In {\em 2009 IEEE 12th international conference on computer vision},
  pages 2106--2113. IEEE, 2009.

\bibitem{Kienzle}
Wolf Kienzle, Felix~A Wichmann, Matthias~O Franz, and Bernhard Sch{\"o}lkopf.
\newblock A nonparametric approach to bottom-up visual saliency.
\newblock In {\em Advances in neural information processing systems}, pages
  689--696, 2007.

\bibitem{koch1987shifts}
Christof Koch and Shimon Ullman.
\newblock Shifts in selective visual attention: towards the underlying neural
  circuitry.
\newblock In {\em Matters of intelligence}, pages 115--141. Springer, 1987.

\bibitem{AlexNet}
Alex Krizhevsky, Ilya Sutskever, and Geoffrey~E Hinton.
\newblock Imagenet classification with deep convolutional neural networks.
\newblock In {\em Advances in neural information processing systems}, pages
  1097--1105, 2012.

\bibitem{MSI-NET}
Alexander Kroner, Mario Senden, Kurt Driessens, and Rainer Goebel.
\newblock Contextual encoder-decoder network for visual saliency prediction.
\newblock {\em Neural Networks}, 2020.

\bibitem{mit-tuebingen-saliency-benchmark}
Matthias K{\"u}mmerer, Zoya Bylinskii, Tilke Judd, Ali Borji, Laurent Itti,
  Fr{\'e}do Durand, Aude Oliva, and Antonio Torralba.
\newblock Mit/tübingen saliency benchmark.
\newblock https://saliency.tuebingen.ai/, 2019.

\bibitem{deepgazei}
Matthias K{\"u}mmerer, Lucas Theis, and Matthias Bethge.
\newblock Deep gaze i: Boosting saliency prediction with feature maps trained
  on imagenet.
\newblock {\em arXiv preprint arXiv:1411.1045}, 2014.

\bibitem{IGain}
Matthias K{\"u}mmerer, Thomas~SA Wallis, and Matthias Bethge.
\newblock Information-theoretic model comparison unifies saliency metrics.
\newblock {\em Proceedings of the National Academy of Sciences},
  112(52):16054--16059, 2015.

\bibitem{kummererSaliencyBenchmarkingMade2018}
Matthias K{\"u}mmerer, Thomas S.~A. Wallis, and Matthias Bethge.
\newblock Saliency benchmarking made easy: Separating models, maps and metrics.
\newblock In Vittorio Ferrari, Martial Hebert, Cristian Sminchisescu, and Yair
  Weiss, editors, {\em Computer Vision – {ECCV} 2018}, Lecture Notes in
  Computer Science, pages 798--814. Springer International Publishing, 2018.

\bibitem{kummererUnderstandingLowHighLevel2017}
Matthias K{\"u}mmerer, Thomas S.~A. Wallis, Leon~A. Gatys, and Matthias Bethge.
\newblock Understanding {{Low}}- and {{High}}-{{Level Contributions}} to
  {{Fixation Prediction}}.
\newblock In {\em The {{IEEE International Conference}} on {{Computer Vision}}
  ({{ICCV}})}, pages 4789--4798, 2017.

\bibitem{PASCALS}
Yin Li, Xiaodi Hou, Christof Koch, James~M Rehg, and Alan~L Yuille.
\newblock The secrets of salient object segmentation.
\newblock In {\em Proceedings of the IEEE conference on computer vision and
  pattern recognition}, pages 280--287, 2014.

\bibitem{deepfeat}
Ali Mahdi, Jun Qin, and Garth Crosby.
\newblock Deepfeat: A bottom-up and top-down saliency model based on deep
  features of convolutional neural networks.
\newblock {\em IEEE Transactions on Cognitive and Developmental Systems},
  12(1):54--63, 2019.

\bibitem{ovadiaCanYouTrust2019}
Yaniv Ovadia, Emily Fertig, Jie Ren, Zachary Nado, D. Sculley, Sebastian
  Nowozin, Joshua Dillon, Balaji Lakshminarayanan, and Jasper Snoek.
\newblock Can you trust your model's uncertainty? evaluating predictive
  uncertainty under dataset shift.
\newblock {\em Advances in Neural Information Processing Systems}, 32, 2019.

\bibitem{pan2017salgan}
Junting Pan, Cristian~Canton Ferrer, Kevin McGuinness, Noel~E O'Connor, Jordi
  Torres, Elisa Sayrol, and Xavier Giro-i Nieto.
\newblock Salgan: Visual saliency prediction with generative adversarial
  networks.
\newblock {\em arXiv preprint arXiv:1701.01081}, 2017.

\bibitem{contrast2016}
KH Ruddock, DS Wooding, and SK Mannan.
\newblock The relationship between the locations of spatial features and those
  of fixations made during visual examination of briefly presented images.
\newblock {\em Spatial vision}, 10(3):165--188, 1996.

\bibitem{VGG}
Karen Simonyan and Andrew Zisserman.
\newblock Very deep convolutional networks for large-scale image recognition.
\newblock {\em arXiv preprint arXiv:1409.1556}, 2014.

\bibitem{sticklandDiverseEnsemblesImprove2020}
Asa~Cooper Stickland and Iain Murray.
\newblock Diverse ensembles improve calibration.
\newblock {\em {arXiv}:2007.04206 [cs, stat]}, 2020.

\bibitem{efficentNet}
Mingxing Tan and Quoc~V Le.
\newblock Efficientnet: Rethinking model scaling for convolutional neural
  networks.
\newblock {\em arXiv preprint arXiv:1905.11946}, 2019.

\bibitem{Torralba}
Antonio Torralba, Aude Oliva, Monica~S Castelhano, and John~M Henderson.
\newblock Contextual guidance of eye movements and attention in real-world
  scenes: the role of global features in object search.
\newblock {\em Psychological review}, 113(4):766, 2006.

\bibitem{featureIntegration}
Anne~M Treisman and Garry Gelade.
\newblock A feature-integration theory of attention.
\newblock {\em Cognitive psychology}, 12(1):97--136, 1980.

\bibitem{eDN}
Eleonora Vig, Michael Dorr, and David Cox.
\newblock Large-scale optimization of hierarchical features for saliency
  prediction in natural images.
\newblock In {\em Proceedings of the IEEE Conference on Computer Vision and
  Pattern Recognition}, pages 2798--2805, 2014.

\bibitem{yangDilatedInceptionNetwork2019}
Sheng Yang, Guosheng Lin, Qiuping Jiang, and Weisi Lin.
\newblock A {{Dilated Inception Network}} for {{Visual Saliency Prediction}}.
\newblock {\em arXiv:1904.03571 [cs]}, May 2019.

\bibitem{yarbus1967eye}
Alfred~L Yarbus.
\newblock Eye movements during perception of complex objects.
\newblock In {\em Eye movements and vision}, pages 171--211. Springer, 1967.

\bibitem{SUN}
Lingyun Zhang, Matthew~H Tong, Tim~K Marks, Honghao Shan, and Garrison~W
  Cottrell.
\newblock Sun: A bayesian framework for saliency using natural statistics.
\newblock {\em Journal of vision}, 8(7):32--32, 2008.

\end{thebibliography}
}

\end{document}


\section*{DeepGaze IIE: Calibrated prediction in and out-of-domain for state-of-the-art saliency modeling\\Supplementary Material}
\begin{figure*}[htbp]
\begin{center}
\includegraphics[width=0.45\textwidth]{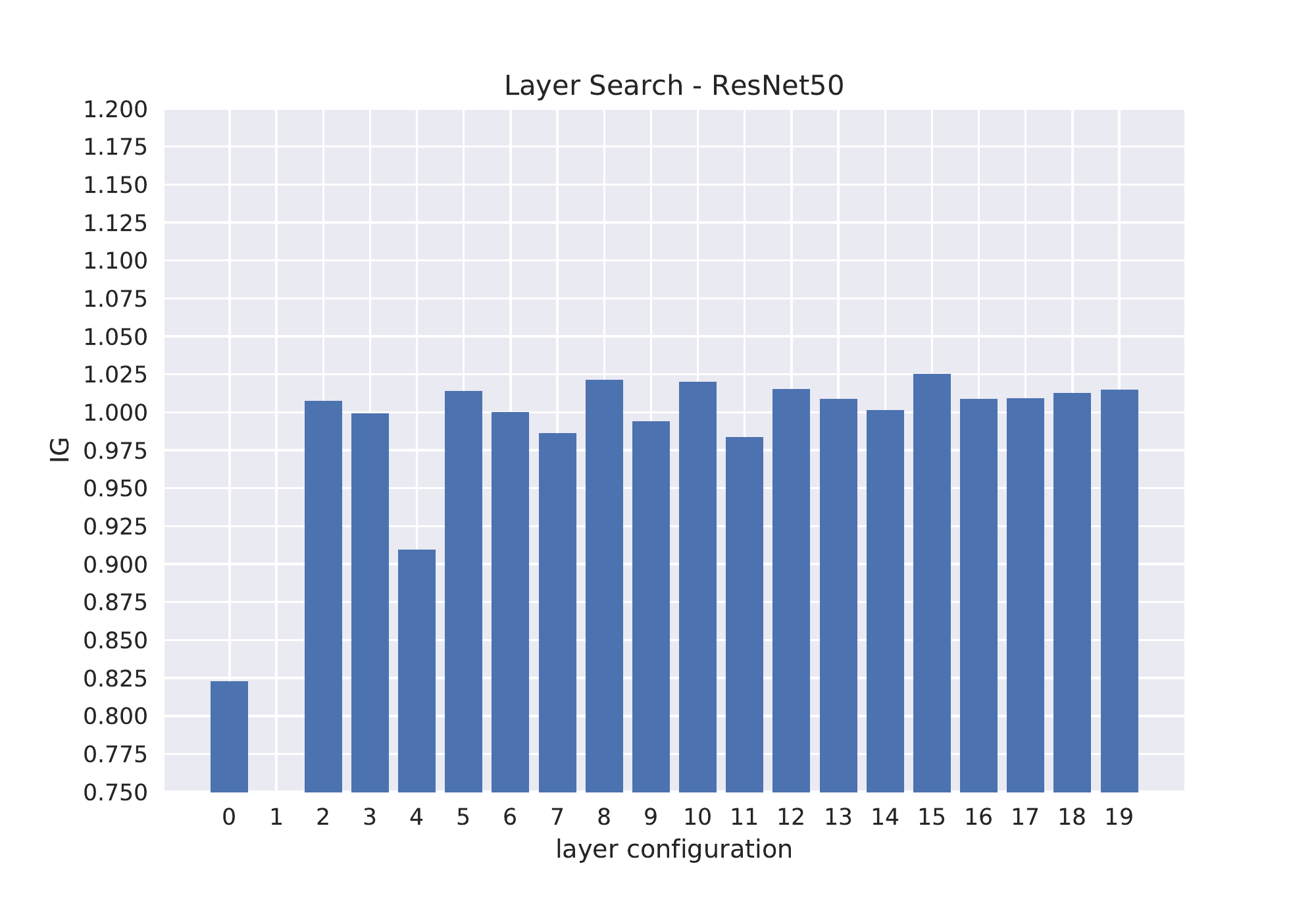}
\includegraphics[width=0.45\textwidth]{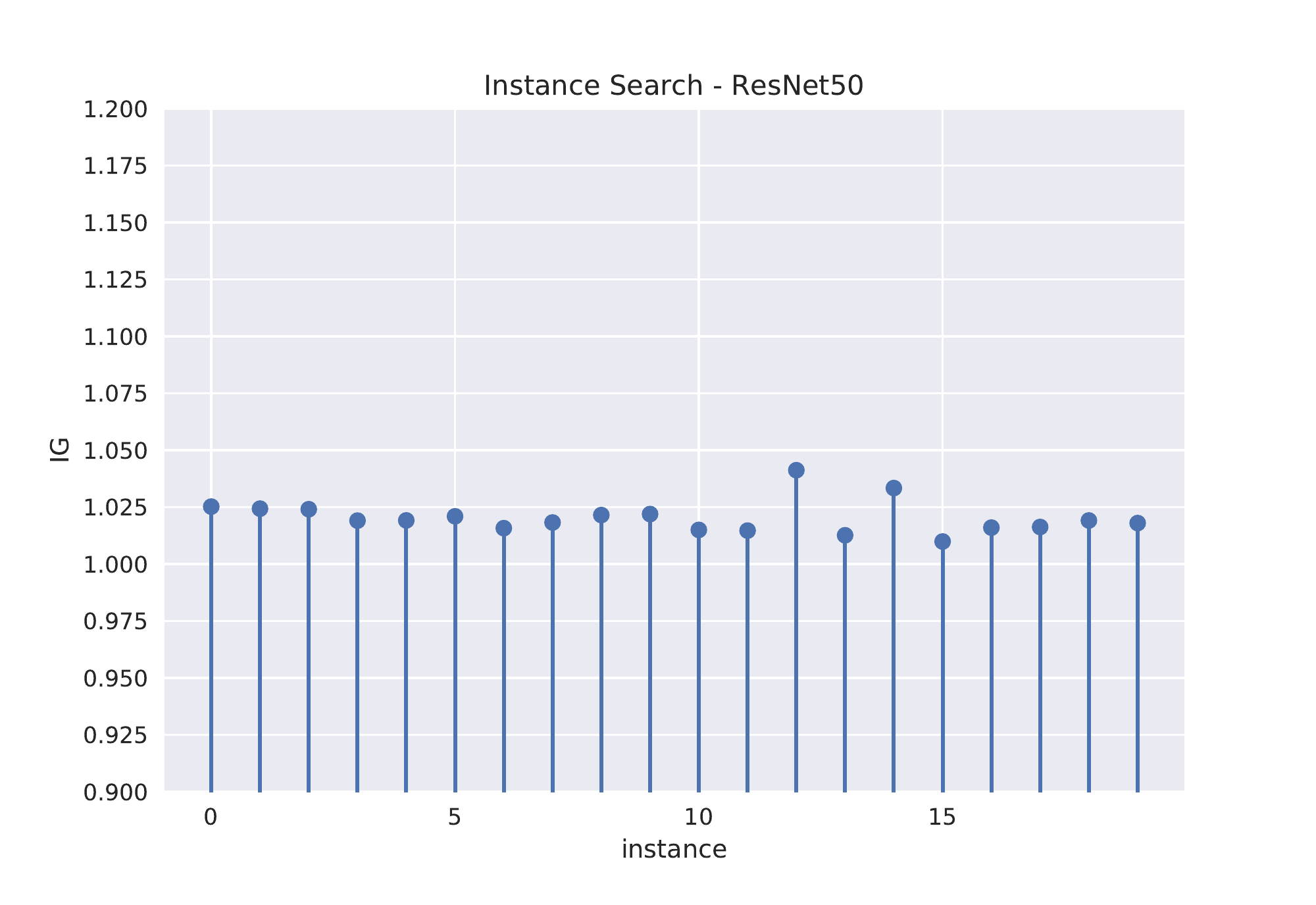}
\caption{ResNet50 layer search (top image) reflects our experiments that involved trying out layers from ResNet50's final convolutional blocks as features. In the case of instance search (bottom image) we simply pick the top performing layer configuration and repeat the same experiment with different seeds (hence different initialization). We can see that the fluctuations between different instances are just as high as the ones we see among the top 5 layer configurations.}
\label{fig:resnet50}
\end{center}
\end{figure*}

\begin{figure*}[t]
\begin{center}
\includegraphics[width=0.48\textwidth]{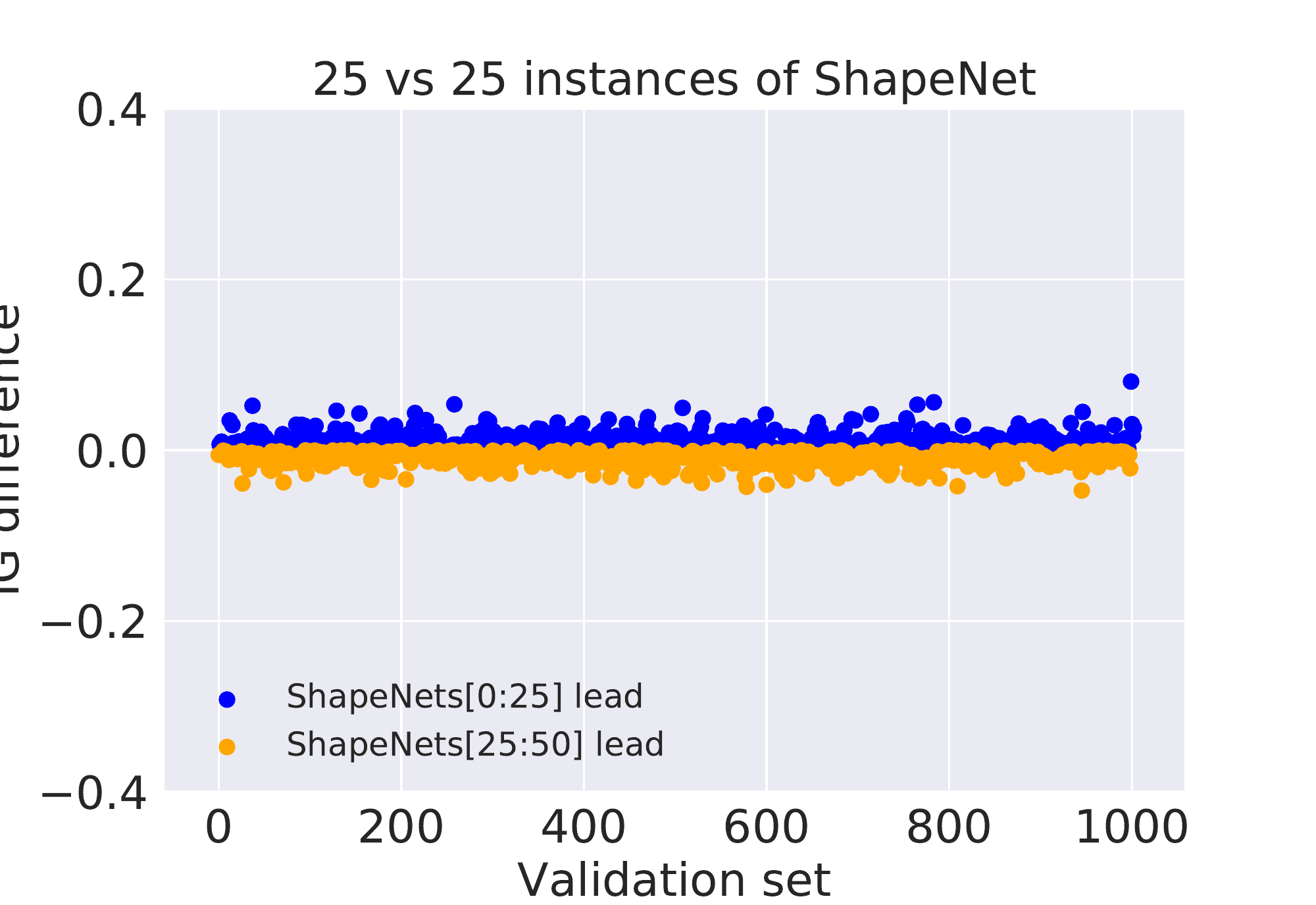}
\includegraphics[width=0.48\textwidth]{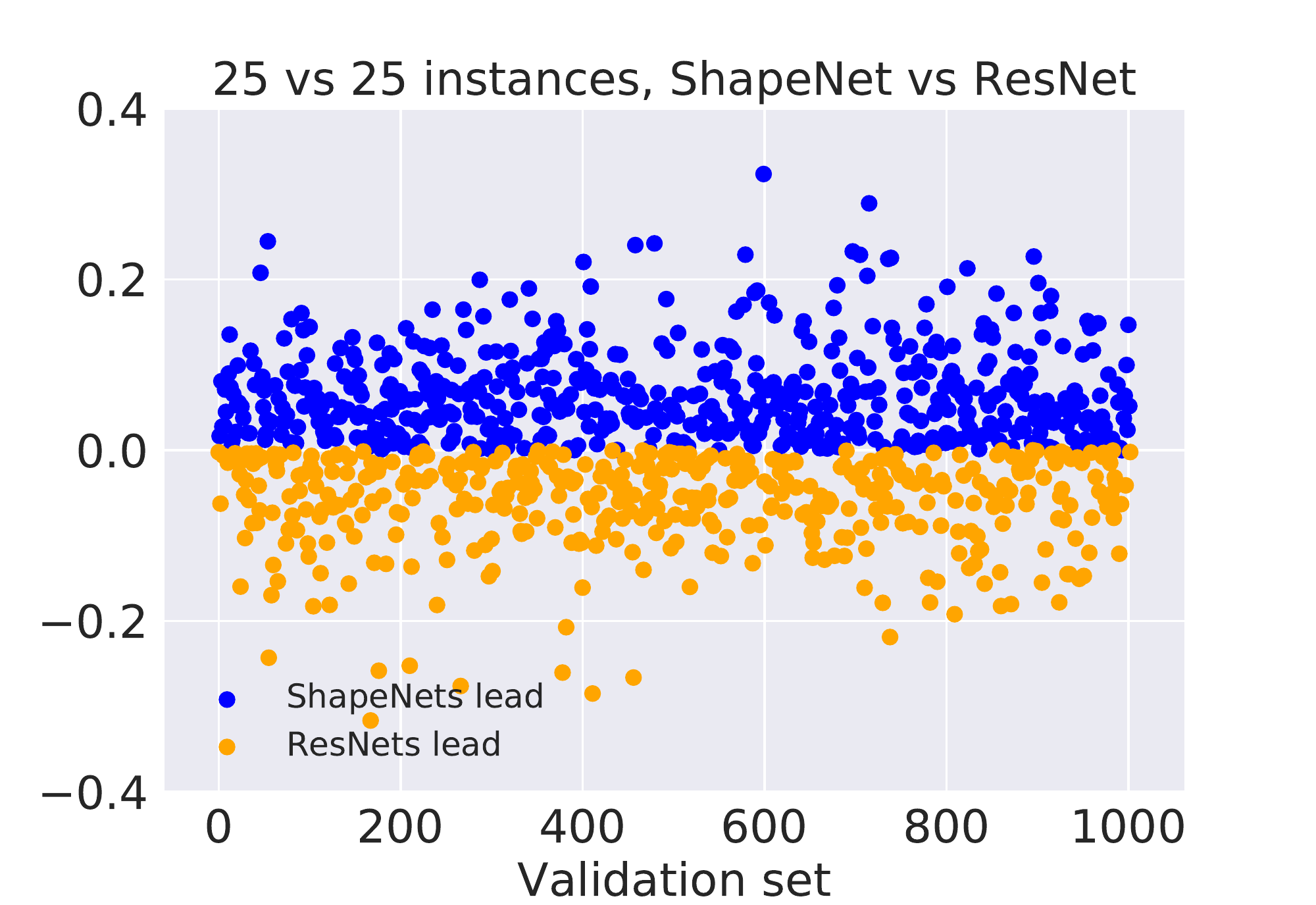}
\caption{Per-image performance variance in between different models. Each point on the X axis corresponds to an image from MIT1003 while the Y axis is the information gain \textit{difference} between the two groups of models, meaning information gain was calculated and averaged across one group of models then subtracted between the two. Thus, the different colors signify which of the two groups is leading in the corresponding sample. 
On the left plot, we compare 50 instances of ShapeNetC in groups of 25 and find that even with the exact same architecture, the standard deviation is a non-marginal value of 0.015. However, when we compared groups of ShapeNetC to ResNet50 (right plot) we found a significant standard deviation of 0.086 in their per-image information gain difference.}
\label{fig:persample-comparison}
\end{center}
\end{figure*}

\begin{figure*}[t]
\begin{center}
\includegraphics[width=0.48\textwidth]{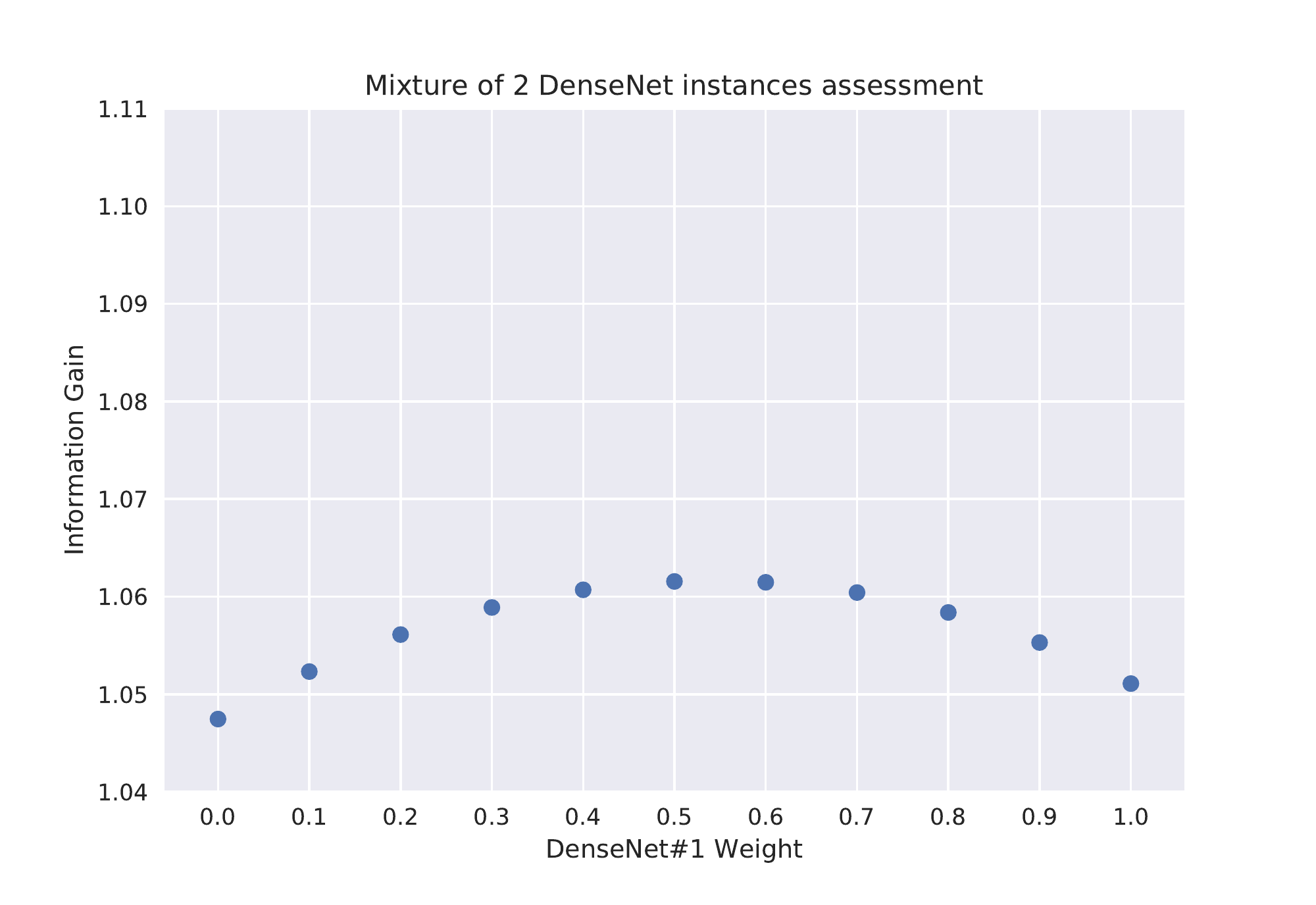}
\includegraphics[width=0.48\textwidth]{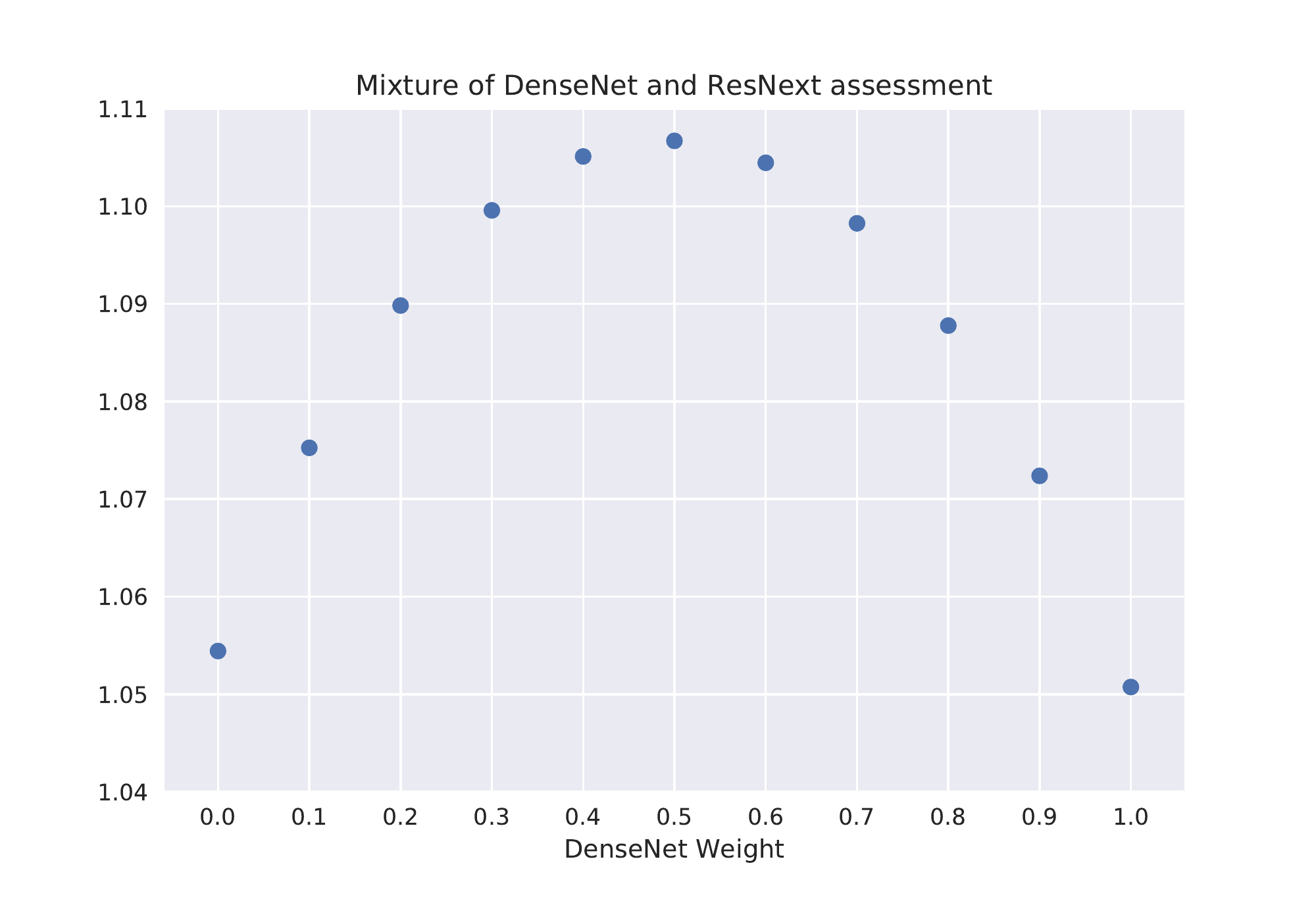}
\caption{Mixtures of models with varying weights. We show the performance when using a mixture of two models with varying mixing coefficients, so that at 0 we see the individual performance of one instance, at 1 that of the other instance while at 0.5 both have equal say at the final density. The left figure shows performances of average densities from instances that use DenseNet-201 as a backbone and is indicative of intra-model complementarity while in the right figure it's instances from two distinct distinct backbones (ResNext50 and DenseNet201) and indicative of inter-model complementarity. Even when mixing instances of the exact same model there is a boost in performance that peaks at the point where each model has equal weight; however, we reach a much higher performance when mixing instances of different models. We empirically found this to be true when combining other models presented in this paper as well.}
\label{fig:mixtures-comparison}
\end{center}
\end{figure*}

\begin{figure*}[htbp]
\begin{center}
\includegraphics[width=\textwidth]{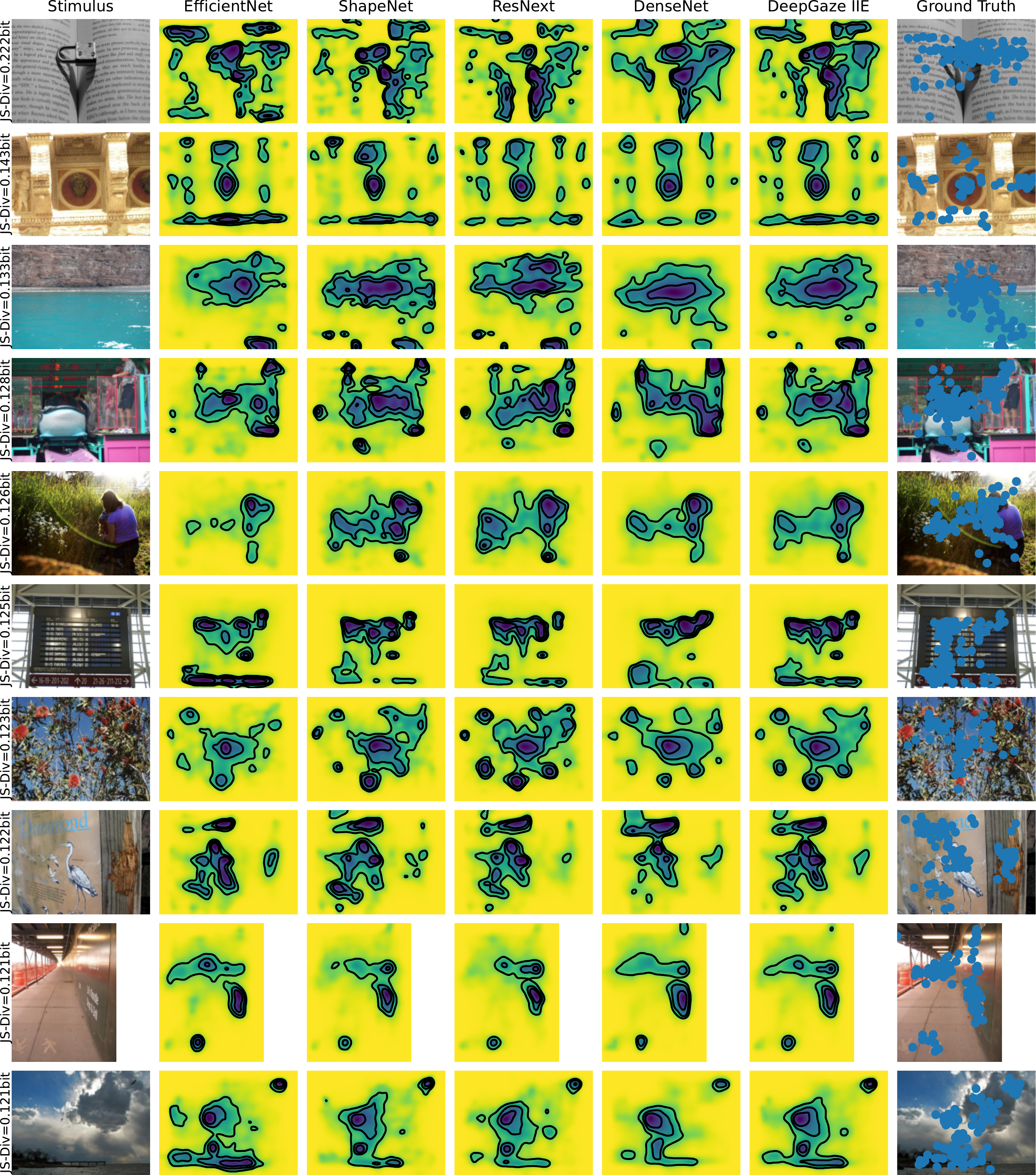}
\caption{We predict the fixation densities from different models using samples of the MIT1003 dataset. To select samples where our models are qualitatively different, we compute the Jensen-Shannon divergence (JS-Div) per image amongst our top four models (not including the mixture DSRE), using mixture of 20 instances per model, thus removing any noise caused by intra-model variability. These are the top-10 images in terms of maximal disagreement and are displayed top to bottom with respect to their JS-Div, the maximum being 0.222 bits corresponding to the top row image.
}
\label{fig:qualitative_analysis}
\end{center}
\end{figure*}

As mentioned in the main text, our initial experiments involved layer selection experiments (10 experiments with 10 different configurations) followed by 20 repeated experiments for initialization selection (figure \ref{fig:resnet50}). We also conducted a detailed evaluation on MIT1003 on all backbones and saliency metrics (table \ref{tab:full_eval}). Our combinatoric experiments were firstly inspired from highlighting how different backbones perform significantly different on a per-sample basis (figure \ref{fig:persample-comparison}). Towards the making of our ensembles, we first tried different weights and found that performance consistently peaks when different backbones get an equal say on the prediction (figure \ref{fig:mixtures-comparison}). Finally, we conducted a principled qualitative analysis using samples whose predictions are maximally different across the backbones of our ensemble (figure \ref{fig:qualitative_analysis}).

\begin{table*}[htbp]
    \centering
    \caption{Evaluation on all models on MIT1003. To assure the robustness of our metrics, we calculate performance in each metric for 20 instances per model, then take the average}
        \begin{tabular*}{\textwidth}{l@{\extracolsep{\fill}}cccccc}
        \toprule
        Backbone &      IG $\uparrow$ &     AUC $\uparrow$ &    sAUC $\uparrow$ &     NSS $\uparrow$ &      CC $\uparrow$ &   KLDiv $\downarrow$ \\
        \midrule
        densenet201  &  \textbf{1.0377} & \textbf{ 0.8892} &  \textbf{0.7876} &  2.5994 &  \textbf{0.7736} &  \textbf{0.5156} \\
        resnext50    &  1.0368 &  0.8886 &  0.7854 &  2.6354 &  0.7731 &  0.5214 \\
        efficientnet &  1.0326 &  0.8890 &  0.7870 &  2.6213 &  0.7704 &  0.5237 \\
        shapenetC    &  1.0278 &  0.8878 &  0.7848 &  \textbf{2.6380} &  0.7716 &  0.5263 \\
        resnet50     &  1.0201 &  0.8874 &  0.7834 &  2.6141 &  0.7657 &  0.5318 \\
        resnet101    &  1.0045 &  0.8866 &  0.7816 &  2.5909 &  0.7631 &  0.5389 \\
        vgg19        &  0.9483 &  0.8838 &  0.7747 &  2.5457 &  0.7486 &  0.5653 \\
        vgg11        &  0.9035 &  0.8803 &  0.7681 &  2.4905 &  0.7346 &  0.5896 \\
        alexnet      &  0.8046 &  0.8736 &  0.7554 &  2.3073 &  0.6983 &  0.6482 \\
        \bottomrule
        \hline
        \end{tabular*}
    \label{tab:full_eval}
\end{table*}

\begin{table*}[]
    \centering
    \caption{Performance of DeepGaze IIE on the SALICON test set. For this version of DeepGaze IIE, we average the individual models after pretraining on the SALICON training dataset, i.e. without finetuning on MIT1003. The SALICON competition does not support proper evaluation of probabilistic models, but only of classic saliency maps. Therefore all reported scores are for saliency maps optimal for NSS (i.e. predicted fixation densities), except for sAUC, for which we used the correct saliency maps for sAUC (i.e., predicted fixation density divided by the average of the predicted fixation densities for all other images).}
    \begin{tabular}{cccccccc}
        \toprule
        Model & sAUC & IG & NSS & CC & AUC & SIM & KL \\
        \midrule
        DeepGaze IIE
          & 0.767 
          & 0.766  
          & 1.996 
          & 0.872 
          & 0.869 
          & 0.733 
          & 0.285 
          \\
        \bottomrule
    \end{tabular}
    \label{tab:my_label}
\end{table*}